\pdfoutput=1
\documentclass[11pt]{article}

\usepackage[final]{acl}

\usepackage{times}
\usepackage{latexsym}
\usepackage[normalem]{ulem}

\usepackage[T1]{fontenc}

\usepackage[utf8]{inputenc}

\usepackage{microtype}
\usepackage{amsmath}
\usepackage{xcolor}
\usepackage{amssymb}
\usepackage{inconsolata}

\usepackage{graphicx}
\usepackage{geometry}

\usepackage{booktabs}
\usepackage{tabularx}
\usepackage{array}

\usepackage{multicol}

\usepackage{tcolorbox}
\usepackage[table]{xcolor}
\tcbuselibrary{breakable, skins}

\definecolor{vRed}{RGB}{200, 30, 30}    
\definecolor{vGreen}{RGB}{34, 139, 34}  
\definecolor{vBlue}{RGB}{30, 70, 180}   

\definecolor{myTabColor}{RGB}{89, 132, 196}  

\newcommand{\imp}[1]{\textbf{\textcolor{vRed}{#1}}}   
\newcommand{\res}[1]{\textbf{\textcolor{vGreen}{#1}}} 
\newcommand{\pau}[1]{\textbf{\textcolor{vBlue}{#1}}}  

\newtcolorbox{DataBox}[1][]{
	enhanced,
	breakable,
	colback=gray!5,
	colframe=gray!40,
	coltitle=black,
	fonttitle=\bfseries,
	title={#1},
	boxrule=0.8pt,
	arc=2pt,
	left=5pt, right=5pt, top=5pt, bottom=5pt
}

\newtcolorbox{ManualBox}[1][]{
	enhanced,
	breakable,             
	colback=gray!5,        
	colframe=gray!50,      
	coltitle=black,        
	fonttitle=\bfseries\large,
	boxrule=1pt,           
	arc=4pt,               
	titlerule=0.5pt,       
	titlerule style=gray,
	title={#1},            
	top=10pt, bottom=10pt, left=10pt, right=10pt 
}

\usepackage{listings}
\usepackage{xcolor}

\definecolor{codegreen}{rgb}{0,0.6,0}
\definecolor{codegray}{rgb}{0.5,0.5,0.5}
\definecolor{codepurple}{rgb}{0.58,0,0.82}
\definecolor{backcolour}{rgb}{0.95,0.95,0.92}

\lstdefinelanguage{json}{
	basicstyle=\ttfamily\small,
	numbers=none,
	numbersep=5pt,
	backgroundcolor=\color{backcolour},
	showspaces=false,
	showstringspaces=false,
	showtabs=false,
	tabsize=2,
	captionpos=b,
	breaklines=true,
	breakatwhitespace=true,
	frame=single,
	commentstyle=\color{codegreen},
	keywordstyle=\color{magenta},
	stringstyle=\color{codepurple},
	morestring=[b]",
	literate=
	*{0}{{{\color{black}0}}}{1}
	{1}{{{\color{black}1}}}{1}
	{2}{{{\color{black}2}}}{1}
	{3}{{{\color{black}3}}}{1}
	{4}{{{\color{black}4}}}{1}
	{5}{{{\color{black}5}}}{1}
	{6}{{{\color{black}6}}}{1}
	{7}{{{\color{black}7}}}{1}
	{8}{{{\color{black}8}}}{1}
	{9}{{{\color{black}9}}}{1}
}

\hypersetup{pdfborder={0 0 0}}

\title{LitVISTA: A Benchmark for Narrative Orchestration in Literary Text}

\author{
\textbf{Mingzhe Lu}$^{1,2}$, 
\textbf{Yiwen Wang}$^{3}$, 
\textbf{Yanbing Liu}$^{1,2}$\thanks{Corresponding authors}, 
\textbf{Qi You}$^{1,2}$, 
\textbf{Chong Liu}$^{3}$, \\
\textbf{Ruize Qin}$^{4}$, 
\textbf{Haoyu Dong}$^{1,2}$, 
\textbf{Wenyu Zhang}$^{3}$, 
\textbf{Jiarui Zhang}$^{1,2}$, 
\textbf{Yue Hu}$^{1,2}$, 
\textbf{Yunpeng Li}$^{1,2*}$ \\
$^{1}$Institute of Information Engineering, Chinese Academy of Sciences, Beijing, China \\
$^{2}$School of Cyber Security, University of Chinese Academy of Sciences, Beijing, China \\
$^{3}$University of Science and Technology of China \quad $^{4}$University of Melbourne \\
\texttt{\{liuyanbing, liyunpeng\}@iie.ac.cn} \\
\includegraphics[width=0.5cm]{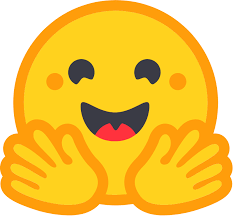} \href{https://huggingface.co/datasets/VivldArc/VISTA}{https://huggingface.co/datasets/VivldArc/VISTA}
}

\begin{document}
\maketitle
\begin{abstract}
Computational narrative analysis aims to capture rhythm, tension, and emotional dynamics in literary texts. Existing large language models can generate long stories but overly focus on causal coherence, neglecting the complex story arcs and orchestration inherent in human narratives. This suggests a structural misalignment between model- and human-generated narratives.
We therefore position narrative analysis as a diagnostic proxy for generation and propose VISTA Space, a high-dimensional framework for narrative orchestration that unifies human and model perspectives while jointly characterizing narrative function and structure in a common space.
We further introduce LitVISTA, a structurally annotated benchmark grounded in literary texts, which operationalizes VISTA Space for systematic evaluation of models’ narrative orchestration capabilities. 
Under an oracle setting with gold event anchors, we evaluate frontier LLMs including GPT, Claude, Grok, and Gemini. Results reveal systematic deficiencies, as current models struggle to jointly capture narrative function and structure and fail to form an integrated global view of literary narrative orchestration. End-to-end analysis further shows that failures are dominated by anchor identification and localization errors. Even advanced thinking modes yield mixed and often limited gains for literary narrative understanding.
\end{abstract}

\section{Introduction}
Computational narrative analysis lies at the intersection of natural language processing and literary studies, aiming to represent the complex phenomena of storytelling in structured, analyzable forms~\cite{DBLP:series/synthesis/2013Mani,DBLP:conf/aaaifs/LakoffN10,Bal1986NarratologyIT}. While human meaning-making is articulated through language, in literary narratives, this articulation goes beyond simple action sequences~\cite{Bruner1991TheNC,Herman2009BasicEO}. Authors deliberately orchestrate events to externalize perceptions, intentions, and mental states, creating a specific rhythm of experience~\cite{Zunshine2006WhyWR,Genette1980NarrativeD}. Accordingly, narrative events are not functional equivalents; they are organized to serve distinct structural roles~\cite{Barthes1975AnIT,Chatman1979StoryAD}. Capturing these differences is central to modeling the pacing and tension~\cite{Brewer1982StoriesAT} that distinguish compelling literature from mere coherence.

\begin{figure}[t]
	\includegraphics[width=0.95\linewidth]{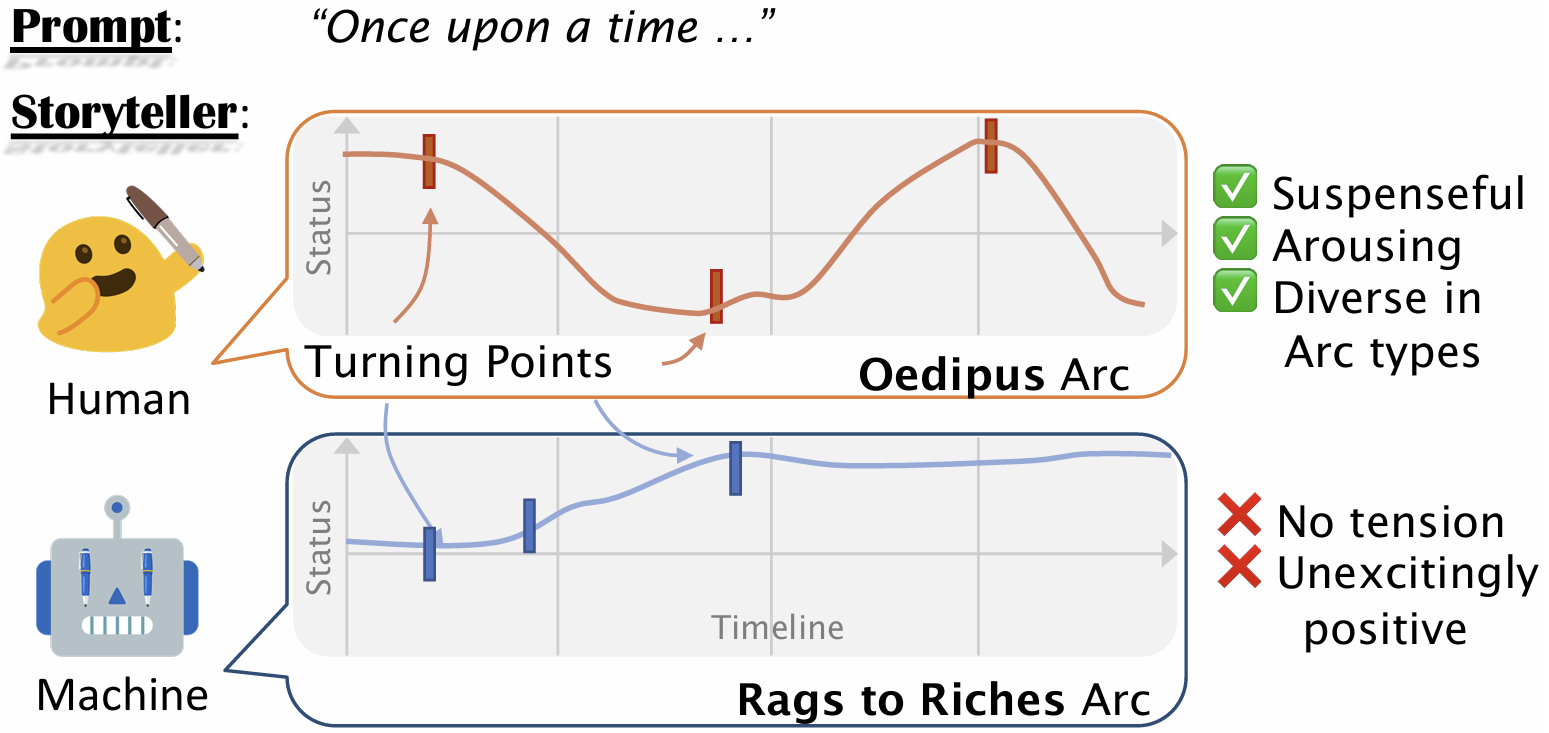}
	\caption{Comparison of story arcs between human and LLM storytellers. This image, reproduced from~\cite{tian-etal-2024-large-language}, shows that LLM-generated stories often have simpler arcs and earlier turning points, whereas human-authored narratives are more complex.}
	\label{intro_comparsion}
\end{figure}

\begin{figure*}[t]
	\centering
	\includegraphics[width=0.95\textwidth]{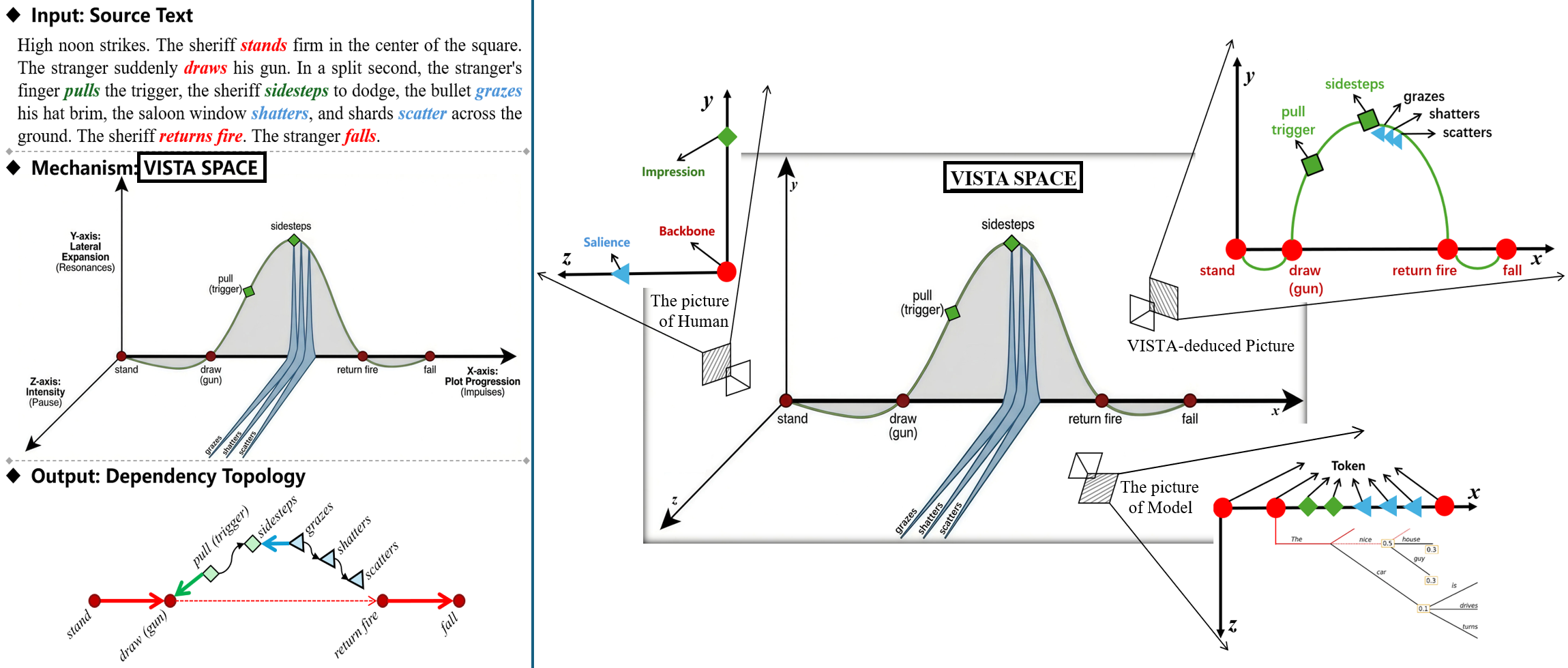}
	\caption{\textbf{VISTA Space and its projections.} The center illustrates VISTA Space, a higher-dimensional representation of narrative orchestration. The surrounding panels show three projections: the human picture of narrative experience (left), the LLM picture based on token-level representations (bottom-right), and the VISTA-induced picture (top-right), which situates human and model representations within a unified structural perspective.}
	\label{fig:intro-example}
\end{figure*}

Existing approaches primarily focus on extending story length while preserving logical consistency~\cite{DBLP:journals/corr/abs-2503-23512,DBLP:conf/pakdd/ParkYJ24,DBLP:conf/cikm/XiaPQ0L0W25}, but such expansion in scale does not yield a commensurate improvement in the actual reading experience. Recent empirical studies~\cite{tian-etal-2024-large-language,DBLP:conf/acl/Wang0H025} reveal systematic differences between human and model narratives at the level of global story shape. As shown in Figure~\ref{intro_comparsion}, the vertical axis tracks the protagonist's fortune from bad to good, and the horizontal axis tracks narrative progression from beginning to end. Human-authored stories are distributed across more diverse arc types, whereas model-generated narratives cluster around more uniformly positive and less inflected trajectories, with turning points tending to occur earlier in the story. This pattern suggests that current models still underdevelop major setbacks and climactic progression, yielding flatter pacing and weaker suspense than human-authored narratives.

Observations of human reading experience suggest that, after reading, readers do not retain the full textual surface of a story, but instead compress it into a mental picture that preserves the narrative backbone, overall atmosphere, and moments of heightened intensity~\cite{Dijk1983StrategiesOD}. This resonates with \textit{Wittgenstein’s picture theory of meaning (Prop. 2.1, 4.01)}~\cite{Wittgenstein2014TractatusL}, according to which understanding consists in forming internal pictures of facts. Computational models likewise develop internal pictures of stories during understanding and generation through the accumulation of probabilistic signals over text. Although both humans and models form such representational surfaces, the principles governing how these pictures are formed differ, giving rise to a structural misalignment between human narrative experiences and model representations.

To bridge this gap, we introduce VISTA (Viewpoint-Integrated Structural Topology for Analysis) Space, a higher-dimensional framework that situates human and model story pictures in a unified space. Within this space, narrative structure becomes an observable object, and event organization is accessed through a dedicated structural plane. This plane captures how narrative dynamics arise from event arrangement, enabling pacing and tension to be visualized, modeled, and measured, while revealing their effects across human and model representations. These representations must be grounded in concrete, annotatable narrative data~\cite{DBLP:books/daglib/0031957} to be empirically accessible.

Importantly, VISTA makes explicit the structural organization that connects narrative production and narrative reception: events are orchestrated to shape pacing, tension, and arc, and these dynamics are in turn reconstructed into a coherent narrative picture. In this work, we position VISTA primarily as a framework for narrative analysis, focusing on recovering narrative topology from a reader-oriented perspective. By rendering narrative orchestration measurable, LitVISTA provides a diagnostic benchmark that reveals where models fail to capture human narrative structure, thereby offering interpretable signals for improving story generation.

Building on this framework, we introduce LitVISTA, a structurally annotated benchmark that makes narrative orchestration explicit in literary texts. LitVISTA represents stories as structured topologies rather than flat sequences, encoding narrative event functions and global dependency relations. Figure~\ref{fig:intro-example} illustrates how a literary passage is mapped into VISTA Space, yielding a VISTA-induced dependency topology. To this end, LitVISTA treats Verbs$^+$ as minimal narrative anchors, covering canonical verbs and event-denoting nominals, and annotates their roles in propagating narrative structure in a signal-like manner, as manifested in forward progression, lateral expansion, and intensity accumulation. As a result, LitVISTA enables systematic evaluation of models’ ability to recover narrative orchestration across events within VISTA Space.

The contributions of this paper can be summarized as follows:
\begin{itemize}
	\item We propose VISTA Space, a higher-dimensional representational framework that conceptualizes literary narrative understanding as the orchestration of events across structural dimensions, providing a unified view of human and model narrative representations.
	\item We introduce LitVISTA, a structurally annotated benchmark grounded in literary texts, which operationalizes VISTA Space for empirical evaluation by mapping narratives into structured event topologies.
	\item Through extensive analysis and evaluation on LitVISTA, we examine the narrative understanding capabilities of existing models, revealing systematic gaps in their ability to orchestrate narrative dynamics.
\end{itemize}

\section{VISTA SPACE}
\subsection{Narrative Proxy}
Human meaning-making is inherently abstract, yet it is expressed through language. In narrative discourse, meaning does not arise from isolated expressions, but from structured configurations that unfold across events. Text therefore serves as the primary medium through which abstract narrative structure is externalized and made observable~\cite{Genette1980NarrativeD}. A key step in modeling narrative organization is thus to identify concrete textual anchors that can reliably proxy such structure~\cite{chambers-jurafsky-2008-unsupervised}.

These anchors must be minimal and well-defined, while remaining representative of underlying narrative dynamics. Verbs naturally fulfill this role as primary carriers of action and change, providing a compact interface between textual form and narrative progression~\cite{Davidson2001TheLF}. 

To support narrative analysis, we extend the notion of verbs beyond strictly grammatical definitions by including event-denoting nominals such as \textit{marriage} and \textit{departure}. These nominal forms retain the argument structure and event semantics of their verbal counterparts~\cite{DBLP:conf/ndqa/PustejovskyCISGSKR03}, allowing them to participate in event representations in a manner similar to verbs.

\paragraph{Terminological Distinction.} Throughout this work, we use the term \textit{Verb}$^+$ to denote a broader class of event anchors. We explicitly distinguish narrative events as abstract units of meaning from Verbs$^+$ as their concrete textual anchors used for computational modeling.

\subsection{Narrative Configuration}
\label{narrative_configuration}
Narrative meaning is not fully captured by the sum of discrete Verbs$^+$; rather, it arises from their configuration across the text~\cite{Polkinghorne2010NarrativeKA}. While a list of Verbs$^+$ can describe what happened, it does not account for how information is structured and presented over time, including the ordering, emphasis, and contextual dependencies among events. The essence of narrative, therefore, lies not in the isolated presence of Verbs$^+$, but in their contribution to the structural architecture.

Within the narrative architecture, different Verbs$^+$ assume distinct structural functions. In practice, the same Verbs$^+$ describing the same situation at the same textual position may be assigned different structural roles within different narrative orchestrations~\cite{Chatman1979StoryAD}, with concrete illustrations provided in Appendix~\ref{app:narrative_practice}. These dynamic role assignments go beyond causality. They allow narrative organization to vary independently of action, giving rise to global properties such as pacing, tension, and rhythm~\cite{Sternberg1992TellingIT}.

\subsection{Narrative Computation}
To implement this structural architecture, we introduce VISTA Space as a computational topology. A key distinction is made between discrete chronological progression and continuous lateral expansion.

Two variables are introduced to represent these dimensions: a discrete \textit{Narrative Progress Index} ($\tau$) that indexes story stages, and a continuous \textit{Marginal Increment} ($\delta$) that measures descriptive expansion without advancing the stage.

\textbf{Definition 1 (Metric Domains).} \textit{The narrative coordinate space is formally constrained by the following domains:}
\begin{equation}
	\label{eq:metric_domains}
	\tau \in \mathbb{N}, \quad \delta \in (0, 1) \subset \mathbb{R}.
\end{equation}

Narrative discourse reconfigures underlying events, distinct from a flat chronology. To capture this structure, we define the \textit{orchestration topology} through a functional mapping that determines how an anchor operates on the narrative state.

\textbf{Definition 2 (Anchor Topology).} \textit{Let $E_\tau$ denote the narrative state at progress index $\tau$. The transition logic $\mathcal{F}(v)$ defines the operation of an anchor on this state:}
\begin{equation}
	\label{eq:anchor_topology}
	\mathcal{F}(v) = \left\{
	\begin{array}{ll}
		E_\tau \to E_{\tau+1},       \\  
		E_\tau \to E_{\tau+\delta},  \\  
		E_\tau \to E_\tau.
	\end{array}
	\right.
\end{equation}

This transition logic establishes a three-dimensional narrative space constructed by three primary functional roles, with a residual category for syntactic elements:

\textbf{Impulses ($\mathcal{V}_{I}$):} Anchors where $\mathcal{F}(v): E_\tau \to E_{\tau+1}$. These form the narrative backbone (the X-axis), advancing the plot to a new stage.

\textbf{Resonances ($\mathcal{V}_{R}$):} Anchors where $\mathcal{F}(v): E_\tau \to E_{\tau+\delta}$. These form the enveloping texture (the Y-axis), expanding descriptively without advancing the stage.

\textbf{Pauses ($\mathcal{V}_{P}$):} Anchors where $\mathcal{F}(v): E_\tau \to E_\tau$. These generate vertical intensity (the Z-axis), inducing temporal suspension to maximize the expressive density of the current moment.

\textbf{Non-Events ($\mathcal{V}_{\emptyset}$):} Syntactic elements that do not contribute to the topology.

\textbf{Definition 3 (Narrative Dependency).} \textit{The narrative topology is a directed graph $G=(\mathcal{V}, \mathcal{E})$. The set of valid edges $\mathcal{E}$ is the union of two hierarchical layers:}
\begin{equation}
	\label{eq:narrative_dependency}
	\mathcal{E} \;\subseteq\; \underbrace{(\mathcal{V}_R \times \mathcal{V}_I)}_{\text{Primary Layer}} \;\cup\; \underbrace{(\mathcal{V}_P \times (\mathcal{V}_I \cup \mathcal{V}_R))}_{\text{Recursive Layer}}.
\end{equation}

This formation dictates that Resonances must attach directly to the Backbone ($\mathcal{V}_I$), whereas Pauses may attach recursively to existing structures ($v_P \to v_R \to v_I$). For benchmark operationalization, local non-backbone material may be serialized through intermediate heads for convenience, while preserving ultimate dependence on the same governing backbone state.

\textbf{Definition 4 (VISTA Space).} 
\textit{The VISTA Space is a three-dimensional narrative orchestration space, with its projection planes representing human, model, and computational perspectives.}

As shown in Figure~\ref{fig:intro-example}, we map $\mathcal{V}_{I}$, $\mathcal{V}_{R}$, and $\mathcal{V}_{P}$ into this 3D coordinate system. The X-axis represents the narrative backbone, driven by $\mathcal{V}_{I}$ and quantified by the index $\tau$. The Y-axis characterizes $\mathcal{V}_{R}$, which emerges around $\mathcal{V}_{I}$ and is quantified by $N\delta$, where $N$ denotes the number of $\mathcal{V}_{P}$ elements along the Z-axis that correspond to the current $\mathcal{V}_{R}$. The Z-axis is dedicated to $\mathcal{V}_{P}$, functioning as a unit impulse with amplitude 1, signifying the discrete presence of a pause.

While it might seem intuitive to merge the Z-axis with the Y-axis, as both capture aspects of narrative progression, it is important to note that the VISTA Space is derived from the orthogonal projections of human and model representations. As shown in the left panel of Figure~\ref{fig:intro-example}, these projections are distinct in the human narrative picture. Consequently, modeling the Z-axis is indispensable for capturing this distinct behavioral feature.

\begin{figure*}[t]
	\centering
	\includegraphics[width=0.98\textwidth]{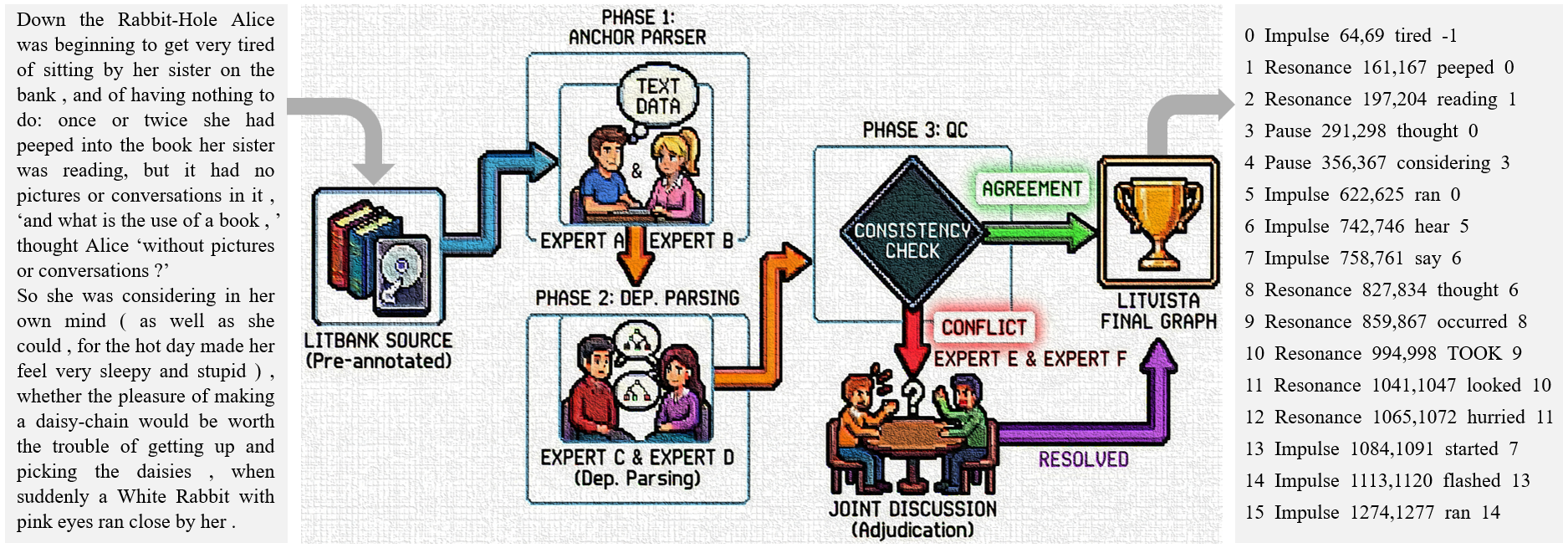}
	\caption{The process begins with LitBank text data. Experts A and B independently annotate Verb$^+$ roles in Phase 1. In Phase 2, dependency parsing is conducted by Experts C and D independently. Phase 3 resolves any conflicts through adjudication, producing the final LitVISTA graph.}
	\label{fig:pipeline}
\end{figure*}

\section{LitVISTA}
In this section, we formally introduce \textbf{LitVISTA}, a structurally annotated benchmark for evaluating and diagnosing models’ narrative orchestration capabilities in literary texts.

\subsection{Dataset Construction}
To ensure rigorous corpus quality, we constructed LitVISTA based on the \textit{LitBank}~\cite{DBLP:conf/lrec/BammanLM20} corpus. 

We adopt LitBank because it provides a curated literary corpus and an established event-centric annotation layer that closely matches our Verb$^+$ notion, covering both verbal and event-denoting nominal anchors. This event layer can be treated as a fixed upstream component in realistic pipelines, allowing LitVISTA to focus on higher-level narrative structure.

The dataset consists of complete narrative chapters, enabling unconstrained long-range topological structure with interleaved $\mathcal{V}_I$, $\mathcal{V}_R$, and recursive $\mathcal{V}_P$ attachments to assess holistic event integration capabilities. Given this design focus, LitVISTA is intentionally positioned as a high-precision evaluation benchmark for narrative orchestration rather than a large-scale training corpus.

\subsection{Annotation Procedure}
\subsubsection{Annotator Background}
\label{anno-background}
The broader development effort involved seven contributors. In the early stage, the team was led by one PhD researcher with interdisciplinary expertise in computational narratology and AI, who was responsible for protocol design, pilot exploration, and guideline refinement. The large-scale annotation was then carried out by six annotators with NLP backgrounds, including three Master's students and three PhD candidates. This division of labor reflects a practical trade-off between narratological grounding and the efficient execution of a technically constrained annotation protocol.

The annotation was conducted in three phases, as illustrated in Figure~\ref{fig:pipeline}. Phases~1 and~2 were each completed by a pair consisting of one PhD annotator and one Master's annotator, whereas Phase~3 was carried out by a different pair, again consisting of one PhD annotator and one Master's annotator. All annotators participated on a voluntary basis and received no monetary compensation.

\subsubsection{Annotation Details}
\label{anno-detail}
In Phase~1, we annotated using the LitBank event triggers as the candidate inventory, which was retained almost entirely. Only a very small number of non-eventive or otherwise non-informative items were removed. The main task in this phase was therefore to assign each retained trigger to its topological category ($\mathcal{V}_I$, $\mathcal{V}_R$, or $\mathcal{V}_P$).

In Phase~2, annotators labeled dependencies for these retained triggers. $\mathcal{V}_I$ nodes form the narrative backbone as a single chain, with the first $\mathcal{V}_I$ assigned index $-1$ to mark the start. $\mathcal{V}_R$ is anchored to the governing $\mathcal{V}_I$ it elaborates, usually locally but sometimes across longer spans, as in flashback or inserted narration. Consecutive $\mathcal{V}_P$ spans are interpreted as co-dependent realizations of the same governing structural state rather than as progressive $\mathcal{V}_P \rightarrow \mathcal{V}_P$ chains. More detailed annotation guidelines are provided in Appendix~\ref{app:manual}.

\subsubsection{Annotation Summary}
Under this protocol, inter-annotator consistency was 0.49 in Phase~1 and 0.76 in Phase~2. We attribute this gap to the openness of literary interpretation: topological role assignment admits multiple plausible readings, whereas dependency relations are usually more explicit. Disagreements were further resolved in Phase~3 to produce the final consensus-derived gold standard, and no inter-annotator score is reported for this stage because it was adjudicative rather than independent. Based on these finalized annotations, we next summarize the resulting dataset statistics.

\begin{table}[h]
	\centering
	\caption{Statistics of the LitVISTA Dataset. Length is measured in tokens.}
	\label{tab:corpus_stats}
	\begin{tabular}{l|ccc} 
		\hline
		Metric & Train & Val & Test \\
		\hline
		Avg. Length        & 10.2k & 9.9k & 10.7k \\
		\hline
		Avg. \# $|V_I|$    & 13.04 & 18.20 & 11.00 \\
		Avg. \# $|V_R|$    & 59.90 & 78.40 & 49.10 \\
		Avg. \# $|V_P|$     & 3.84 & 3.50 & 3.90 \\
		Avg. Cross Dep.      & 75.67 & 100.10 & 63.90 \\
		\hline
	\end{tabular}
\end{table}

Using the final adjudicated annotations, we partitioned the dataset into training, validation, and test sets with an 8:1:1 ratio. Table~\ref{tab:corpus_stats} summarizes the average text length, the distribution of Verb$^+$ subtypes ($\mathcal{V}_I$, $\mathcal{V}_R$, $\mathcal{V}_P$), and the number of cross dependencies in each split. The predominance of $\mathcal{V}_R$ reflects the descriptive emphasis commonly observed in literary narrative discourse, while the frequent cross dependencies further indicate the structural complexity of the annotated narratives.

\begin{table*}[t]
	\centering
	\caption{\textbf{Oracle Evaluation Results on LitVISTA.} 
		We employ a \textbf{heatmap visualization} where color intensity corresponds to performance: \colorbox{myTabColor!50}{\textbf{Darker}} indicates higher scores, and lighter indicates lower scores. Models are sorted by the harmonic mean.}
	\label{tab:main_results}
	\resizebox{\textwidth}{!}{
		\begin{tabular}{l|ccc|ccc|c}
			\hline
			& \multicolumn{6}{c|}{\textit{Oracle Eval}} & \multicolumn{1}{c}{\textit{Overall}} \\
			\hline
			& \multicolumn{3}{c|}{\textbf{Anchor Parsing}} 
			& \multicolumn{3}{c|}{\textbf{Dep. Parsing}} 
			& \textbf{Harmonic} \\ 
			\cline{2-7}
			& P & R & F1 
			& P & R & F1 
			& \textbf{Mean}$\uparrow$\\
			\hline
			GPT-5.1      & \cellcolor{myTabColor!35}0.4066 & \cellcolor{myTabColor!15}0.3393 & \cellcolor{myTabColor!15}0.3033 & \cellcolor{myTabColor!0}0.0746 & \cellcolor{myTabColor!0}0.0464 & \cellcolor{myTabColor!0}0.0460 & \cellcolor{myTabColor!0}0.0799 \\
			GPT-5        & \cellcolor{myTabColor!50}0.4823 & \cellcolor{myTabColor!50}0.4862 & \cellcolor{myTabColor!45}0.4348 & \cellcolor{myTabColor!2}0.1006 & \cellcolor{myTabColor!3}0.1121 & \cellcolor{myTabColor!1}0.0745 & \cellcolor{myTabColor!5}0.1272 \\
			Doubao-seed-1.6-thinking      & \cellcolor{myTabColor!10}0.2914 & \cellcolor{myTabColor!10}0.2956 & \cellcolor{myTabColor!10}0.2890 & \cellcolor{myTabColor!10}0.2066 & \cellcolor{myTabColor!10}0.1772 & \cellcolor{myTabColor!8}0.1456 & \cellcolor{myTabColor!15}0.1936 \\
			Claude-opus-4.5-thinking      & \cellcolor{myTabColor!5}0.2674 & \cellcolor{myTabColor!9}0.2913 & \cellcolor{myTabColor!6}0.2646 & \cellcolor{myTabColor!10}0.2012 & \cellcolor{myTabColor!8}0.1577 & \cellcolor{myTabColor!10}0.1641 & \cellcolor{myTabColor!16}0.2026 \\
			GPT-5.2-pro      & \cellcolor{myTabColor!45}0.4543 & \cellcolor{myTabColor!55}0.5179 & \cellcolor{myTabColor!50}0.4540 & \cellcolor{myTabColor!11}0.2090 & \cellcolor{myTabColor!15}0.2220 & \cellcolor{myTabColor!11}0.1699 & \cellcolor{myTabColor!25}0.2473 \\
			DeepSeek-v3.2-thinking      & \cellcolor{myTabColor!15}0.3123 & \cellcolor{myTabColor!20}0.3440 & \cellcolor{myTabColor!18}0.3140 & \cellcolor{myTabColor!15}0.2564 & \cellcolor{myTabColor!20}0.2799 & \cellcolor{myTabColor!15}0.2219 & \cellcolor{myTabColor!28}0.2600 \\
			ChatGLM-4.7      & \cellcolor{myTabColor!30}0.3708 & \cellcolor{myTabColor!15}0.3225 & \cellcolor{myTabColor!22}0.3362 & \cellcolor{myTabColor!18}0.2890 & \cellcolor{myTabColor!15}0.2314 & \cellcolor{myTabColor!15}0.2182 & \cellcolor{myTabColor!29}0.2646 \\
			Gemini-2.5-pro-thinking  & \cellcolor{myTabColor!16}0.3161 & \cellcolor{myTabColor!30}0.3819 & \cellcolor{myTabColor!16}0.3083 & \cellcolor{myTabColor!19}0.2992 & \cellcolor{myTabColor!25}0.3285 & \cellcolor{myTabColor!20}0.2631 & \cellcolor{myTabColor!32}0.2839 \\
			Grok-4    & \cellcolor{myTabColor!18}0.3297 & \cellcolor{myTabColor!5}0.2619 & \cellcolor{myTabColor!7}0.2669 & \cellcolor{myTabColor!30}0.4185 & \cellcolor{myTabColor!22}0.3057 & \cellcolor{myTabColor!28}0.3365 & \cellcolor{myTabColor!35}0.2977 \\
			GPT-5-thinking      & \cellcolor{myTabColor!0}0.2327 & \cellcolor{myTabColor!0}0.2174 & \cellcolor{myTabColor!0}0.1995 & \cellcolor{myTabColor!50}0.6771 & \cellcolor{myTabColor!50}0.6412 & \cellcolor{myTabColor!50}0.6478 & \cellcolor{myTabColor!36}0.3051 \\
			Claude-sonnet-4.5       & \cellcolor{myTabColor!1}0.2377 & \cellcolor{myTabColor!6}0.2655 & \cellcolor{myTabColor!3}0.2254 & \cellcolor{myTabColor!38}0.4981 & \cellcolor{myTabColor!40}0.5262 & \cellcolor{myTabColor!38}0.4728 & \cellcolor{myTabColor!36}0.3053 \\
			Qwen3-235B-a22      & \cellcolor{myTabColor!11}0.2946 & \cellcolor{myTabColor!22}0.3528 & \cellcolor{myTabColor!12}0.2701 & \cellcolor{myTabColor!25}0.3670 & \cellcolor{myTabColor!30}0.4225 & \cellcolor{myTabColor!29}0.3538 & \cellcolor{myTabColor!36}0.3063 \\
			Gemini-2.5-pro  & \cellcolor{myTabColor!20}0.3360 & \cellcolor{myTabColor!38}0.4178 & \cellcolor{myTabColor!22}0.3346 & \cellcolor{myTabColor!21}0.3162 & \cellcolor{myTabColor!25}0.3562 & \cellcolor{myTabColor!22}0.2911 & \cellcolor{myTabColor!37}0.3113 \\
			Grok-4.1-thinking     & \cellcolor{myTabColor!32}0.3930 & \cellcolor{myTabColor!45}0.4609 & \cellcolor{myTabColor!40}0.4086 & \cellcolor{myTabColor!18}0.2798 & \cellcolor{myTabColor!22}0.3252 & \cellcolor{myTabColor!20}0.2669 & \cellcolor{myTabColor!38}0.3229 \\
			Doubao-seed-1.6      & \cellcolor{myTabColor!9}0.2863 & \cellcolor{myTabColor!7}0.2780 & \cellcolor{myTabColor!9}0.2815 & \cellcolor{myTabColor!39}0.5105 & \cellcolor{myTabColor!38}0.4869 & \cellcolor{myTabColor!37}0.4618 & \cellcolor{myTabColor!40}0.3498 \\
			GPT-5.1-thinking      & \cellcolor{myTabColor!5}0.2662 & \cellcolor{myTabColor!2}0.2458 & \cellcolor{myTabColor!3}0.2410 & \cellcolor{myTabColor!60}0.8135 & \cellcolor{myTabColor!50}0.6441 & \cellcolor{myTabColor!55}0.6799 & \cellcolor{myTabColor!42}0.3559 \\
			Gemini-3-pro-preview-thinking   & \cellcolor{myTabColor!28}0.3619 & \cellcolor{myTabColor!30}0.3879 & \cellcolor{myTabColor!20}0.3285 & \cellcolor{myTabColor!30}0.4209 & \cellcolor{myTabColor!35}0.4674 & \cellcolor{myTabColor!32}0.4061 & \cellcolor{myTabColor!43}0.3632 \\
			Claude-opus-4.5      & \cellcolor{myTabColor!13}0.3058 & \cellcolor{myTabColor!18}0.3368 & \cellcolor{myTabColor!12}0.2947 & \cellcolor{myTabColor!40}0.5147 & \cellcolor{myTabColor!42}0.5627 & \cellcolor{myTabColor!41}0.5083 & \cellcolor{myTabColor!44}0.3731 \\
			GPT-4o              & \cellcolor{myTabColor!16}0.3169 & \cellcolor{myTabColor!4}0.2548 & \cellcolor{myTabColor!5}0.2519 & \cellcolor{myTabColor!58}0.7807 & \cellcolor{myTabColor!58}0.7383 & \cellcolor{myTabColor!58}0.7333 & \cellcolor{myTabColor!44}0.3750 \\
			GPT-5.2      & \cellcolor{myTabColor!38}0.4171 & \cellcolor{myTabColor!48}0.4776 & \cellcolor{myTabColor!38}0.3983 & \cellcolor{myTabColor!28}0.4010 & \cellcolor{myTabColor!29}0.4085 & \cellcolor{myTabColor!29}0.3585 & \cellcolor{myTabColor!45}0.3774 \\
			Claude-sonnet-4.5-thinking       & \cellcolor{myTabColor!19}0.3322 & \cellcolor{myTabColor!32}0.3935 & \cellcolor{myTabColor!21}0.3309 & \cellcolor{myTabColor!36}0.4720 & \cellcolor{myTabColor!39}0.5160 & \cellcolor{myTabColor!37}0.4575 & \cellcolor{myTabColor!46}0.3840 \\
			Gemini-3-pro-preview  & \cellcolor{myTabColor!30}0.3817 & \cellcolor{myTabColor!38}0.4171 & \cellcolor{myTabColor!25}0.3495 & \cellcolor{myTabColor!38}0.4928 & \cellcolor{myTabColor!39}0.5175 & \cellcolor{myTabColor!38}0.4736 & \cellcolor{myTabColor!48}0.4022 \\
			DeepSeek-v3.2       & \cellcolor{myTabColor!14}0.3089 & \cellcolor{myTabColor!20}0.3403 & \cellcolor{myTabColor!16}0.3098 & \cellcolor{myTabColor!48}0.5975 & \cellcolor{myTabColor!48}0.6222 & \cellcolor{myTabColor!47}0.5783 & \cellcolor{myTabColor!48}0.4035 \\
			Claude-opus-4       & \cellcolor{myTabColor!31}0.3868 & \cellcolor{myTabColor!40}0.4284 & \cellcolor{myTabColor!35}0.3779 & \cellcolor{myTabColor!35}0.4603 & \cellcolor{myTabColor!37}0.4923 & \cellcolor{myTabColor!35}0.4414 & \cellcolor{myTabColor!49}0.4072 \\
			Claude-sonnet-4       & \cellcolor{myTabColor!9}0.2893 & \cellcolor{myTabColor!11}0.2987 & \cellcolor{myTabColor!9}0.2838 & \cellcolor{myTabColor!60}\textbf{0.8142} & \cellcolor{myTabColor!60}\textbf{0.8115} & \cellcolor{myTabColor!60}\textbf{0.7968} & \cellcolor{myTabColor!50}0.4185 \\
			Claude-opus-4-thinking       & \cellcolor{myTabColor!34}0.3984 & \cellcolor{myTabColor!42}0.4426 & \cellcolor{myTabColor!38}0.3984 & \cellcolor{myTabColor!40}0.5157 & \cellcolor{myTabColor!40}0.5197 & \cellcolor{myTabColor!38}0.4708 & \cellcolor{myTabColor!52}0.4316 \\
			Claude-sonnet-4-thinking       & \cellcolor{myTabColor!55}\textbf{0.4947} & \cellcolor{myTabColor!60}\textbf{0.5236} & \cellcolor{myTabColor!55}\textbf{0.4914} & \cellcolor{myTabColor!49}0.6104 & \cellcolor{myTabColor!46}0.5981 & \cellcolor{myTabColor!45}0.5624 & \cellcolor{myTabColor!60}\textbf{0.5245} \\
			\hline
		\end{tabular}
	}
\end{table*}

\subsection{LitVISTA Task}
We define the LitVISTA task as a narrative structure reconstruction problem, and evaluate it under an \textit{oracle event-level setting} that requires reconstructing nodes and edges in a single pass. This one-stage formulation mandates the model to capture a global narrative coherence, moving beyond iterative local refinements that often suffer from error propagation.

\subsubsection{Oracle Evaluation}
\label{sec:litvista_task}
We adopt an oracle event-level setting to isolate models’ ability to perform high-level narrative orchestration, under the assumption that candidate event anchors (Verb$^+$) are provided.

Formally, in this oracle setting, the model is provided with the raw text $\mathcal{T}$ along with a set of candidate nodes $\mathcal{V}_{\text{cand}}$ (corresponding to Verb$^+$ tokens). 

The model must simultaneously determine the topological roles for these candidates and resolve their dependencies. This joint optimization is described by the following equations:

\begin{equation}
	\left\{
	\begin{aligned}
		r^* &= \arg \max_{r \in \{\mathcal{V}_I, \mathcal{V}_R, \mathcal{V}_P\}} P(r \mid v, \mathcal{T}), \\
		u^* &= \arg \max_{u \in \mathcal{V}_{\text{cand}} \setminus \{v\}} P(v \to u \mid v, r^*, \mathcal{T}).
	\end{aligned}
	\right.
\end{equation}

where $r^*$ represents the predicted topological role, and $u^*$ represents the predicted parent anchor from the candidates (excluding $v$ itself). This formulation ensures that node classification and dependency resolution are interdependent, reconstructing directed edges that enforce the recursive structure of the narrative.

\subsubsection{Eval Metrics}
Given the discrete node and edge definitions in LitVISTA, we report standard Precision (P), Recall (R), and F1 for both topological role prediction and dependency reconstruction. These metrics provide an operational measure of how well a model recovers the annotated narrative graph at the levels of both node classification and edge prediction. Higher scores indicate closer agreement with the gold-standard topology, although they should be interpreted as benchmark measures rather than exhaustive measures of literary understanding.

\section{Experiments}
\subsection{Evaluation Setup}
We evaluate models' narrative orchestration capabilities on LitVISTA, which renders the VISTA Space computable.

Our main experiments use the oracle event-level setting, where models are evaluated on the provided test set with gold-standard event anchors (trigger + character index). This setting isolates high-level narrative reasoning from upstream extraction errors and enables fair comparison across models. In practical applications, LitVISTA can also be used in a modular pipeline, where a separate event extractor is first applied and its outputs are then passed to LitVISTA for structural prediction; in that case, the resulting scores should be interpreted as reflecting both upstream extraction quality and downstream narrative reasoning.

We consider widely adopted model families, including GPT, Gemini, Grok, and Claude, and compare reasoning-enabled variants with their non-reasoning counterparts. Detailed experimental configurations, including hyperparameter settings and prompt designs, are provided in Appendix~\ref{app:exp_setup}.

We also examine an end-to-end setting in Appendix~\ref{app:end2end}, where models are given only raw text. As shown there, current frontier LLMs fail primarily at anchor identification and span localization, which causes cascading failures before role assignment and dependency prediction can be meaningfully assessed. We therefore do not treat end-to-end scores as the primary measure of narrative orchestration ability, but rather as a diagnostic signal of the interaction between upstream extraction and downstream reasoning.

\subsection{Result Analysis}
We report the performance of all baselines in Table \ref{tab:main_results}, following the oracle evaluation protocol defined in Section~\ref{sec:litvista_task}. To intuitively reveal the underlying trade-offs and behavioral shifts hidden within these numerical comparisons, we further visualize the performance distribution in Figure \ref{fig:oracle_vis}.

\subsubsection{Distribution of Performance}
The heatmap visualization in Table \ref{tab:main_results} provides a clear overview of the overall performance landscape, revealing a pronounced asymmetry between Anchor Parsing and Dependency Parsing across models. Specifically, high performance in one subtask is frequently accompanied by substantially weaker performance in the other, and models that simultaneously achieve strong results on both dimensions are notably scarce. This pattern is most evident in the absence of consistently dark regions across both blocks within the same model row. The same trend is corroborated by the scatter plot in Figure \ref{fig:oracle_vis}, where the upper-right quadrant corresponding to strong performance on both tasks remains largely unpopulated.

\subsubsection{Impact of Thinking}
The connecting lines in Figure~\ref{fig:oracle_vis} show that enabling thinking induces systematic shifts rather than uniform improvements. In some cases, thinking substantially enhances structural modeling. For example, GPT-5.1-thinking exhibits a large performance gain relative to its base counterpart, while simultaneously reducing Anchor accuracy, indicating a redistribution of modeling capacity rather than a consistent improvement.

However, this behavior does not generalize across models. As shown in Table~\ref{tab:main_results}, thinking variants of DeepSeek-v3.2, Claude-opus-4.5, and Gemini-2.5-pro display an overall downward or unstable performance trend when compared with their non-thinking counterparts. Despite isolated improvements in specific configurations, enabling thinking often coincides with broad performance degradation across parsing tasks, suggesting that the induced reasoning process may constrain rather than enrich the model’s representational flexibility.

Taken together, these results indicate that thinking primarily reshapes how models allocate capacity, rather than consistently improving narrative understanding. When narrative modeling is dominated by narrow causal reasoning, gains in localized structure may come at the expense of global event organization. This trade-off is especially limiting for literary narratives, where meaning arises from pacing, tension, figurative relations, and non-linear structure beyond simple causality.

\subsubsection{Family-Specific Patterns}
\begin{figure}[h]
	\centering
	\includegraphics[width=0.94\linewidth]{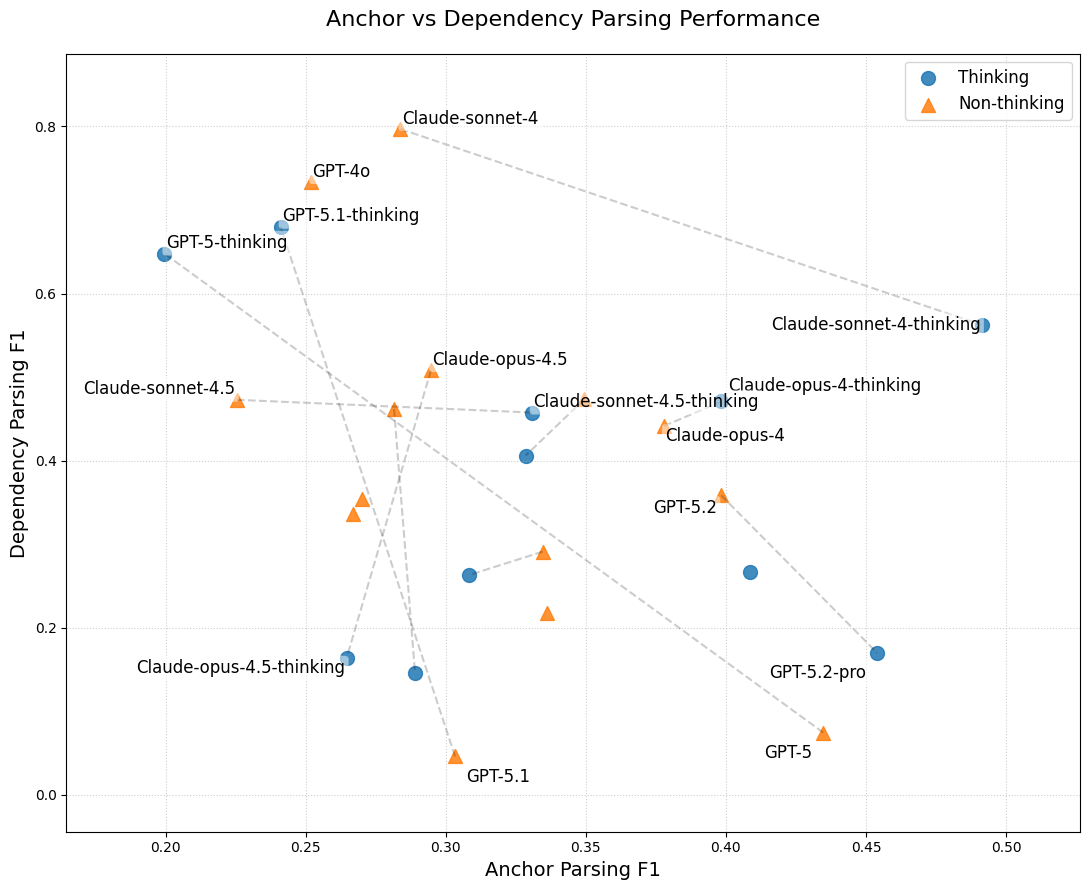}
	\caption{Oracle evaluation results. The scatter plot shows Anchor F1 (x-axis) versus Dependency F1 (y-axis) for each model.}
	\label{fig:oracle_vis}
\end{figure}

While the above analysis already suggests (i) a scarcity of models that are simultaneously strong on both Anchor and Dependency parsing and (ii) non-uniform shifts induced by enabling thinking, these shifts are not arbitrary. Instead, the explicitly labeled models in Figure~\ref{fig:oracle_vis} exhibit family-specific regularities: within the same model family, the thinking-enabled variants tend to move in a more consistent direction, whereas different families display markedly different trajectories.

Concretely, Claude variants largely follow a coherent trend in how thinking reshapes the balance between anchor identification and relational reasoning, while GPT variants exhibit a distinct and often contrasting trend. This divergence indicates that ``thinking'' acts less like a universal improvement knob and more like an amplifier of pre-existing inductive biases encoded by the underlying model family. The connecting lines for the (GPT-5, *-thinking) and (Claude-opus-4.5, *-thinking) pairs appear nearly orthogonal in Figure~\ref{fig:oracle_vis}, a pattern that further underscores this conclusion.

\section{Further Analysis}
In this section, we delve into the unique narrative topologies in LitVISTA to explain why models struggle to comprehend them.

\subsection{Long-Range Narrative Dependencies}
\begin{figure}[h]
	\centering
	\includegraphics[width=0.94\linewidth]{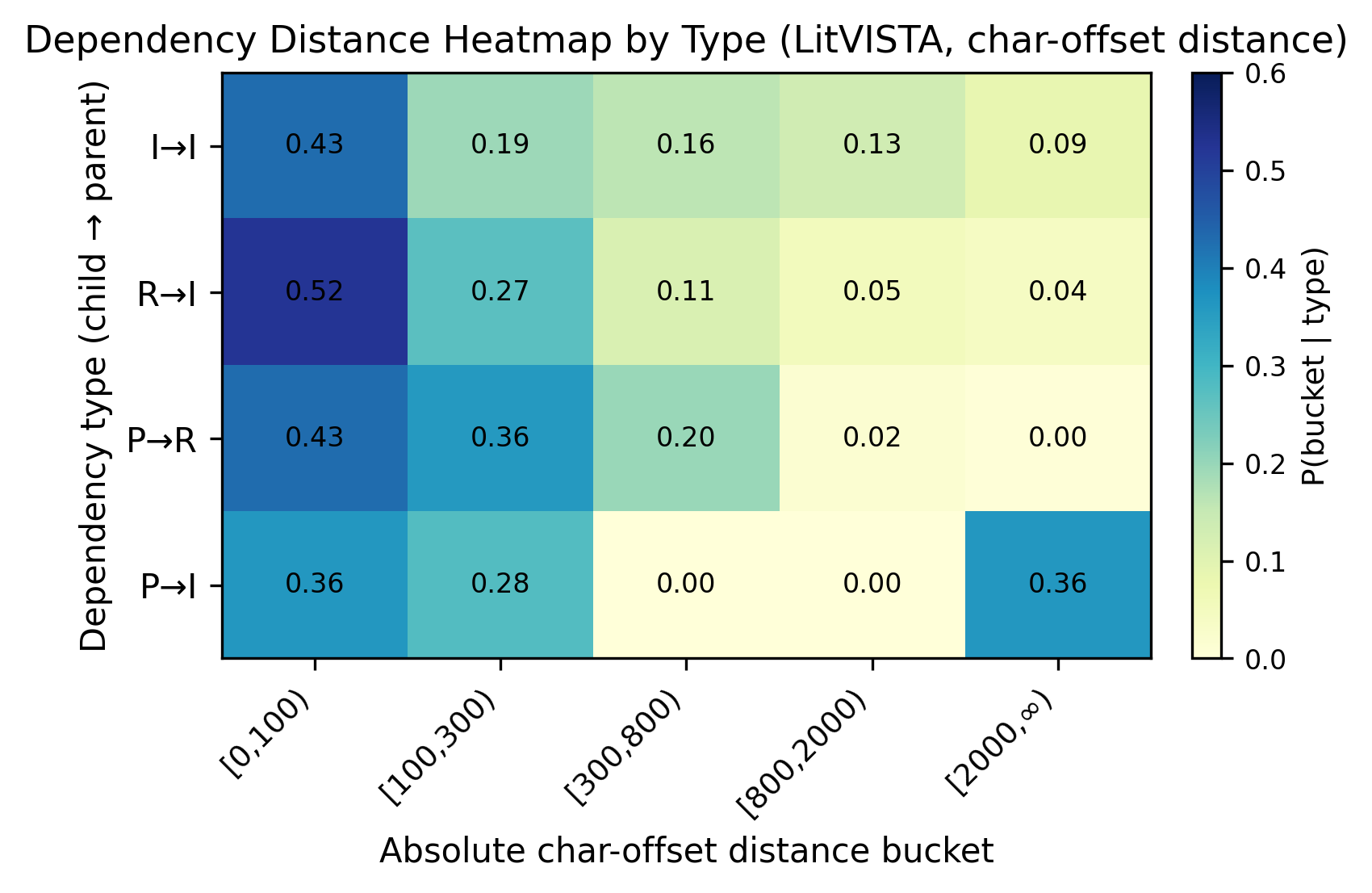}
	\caption{Frequency of narrative dependencies by absolute character offset distance. The X-axis represents distance buckets, and the Y-axis shows different dependency types.}
	\label{fig:dependency_distance}
\end{figure}

As shown in Figure~\ref{fig:dependency_distance}, narrative dependency frequency varies with the absolute textual distance between dependent Verb$^+$ nodes. If such dependencies primarily followed textual proximity, the distribution would concentrate in short-distance intervals.

The observed data, however, exhibit a marked deviation. Although short-range dependencies are common, a substantial proportion, particularly involving Impulse and Pause nodes, spans hundreds or even thousands of characters. Crucially, for several dependency types, long-range associations persist without attenuation.

These findings in dependency patterns suggest that textual proximity is a weak predictor in LitVISTA. \textit{Narrative relations frequently link events that are distant in the linear sequence, because the narrative flow disrupts the timeline or plants foreshadowing}, reflecting higher-level discourse organization. This structural mismatch accounts for the difficulty of understanding, as span-local or next-token-biased models are ill-equipped to capture such non-local topology.

\subsection{Lexical Grounding of Narrative Roles}
Finally, we examine whether narrative roles exhibit lexical regularities by projecting each sufficiently frequent Anchor word’s empirical distribution over Impulse, Resonance, and Pause into a two-dimensional role-preference space.

\begin{figure}[h]
	\centering
	\includegraphics[width=0.93\linewidth]{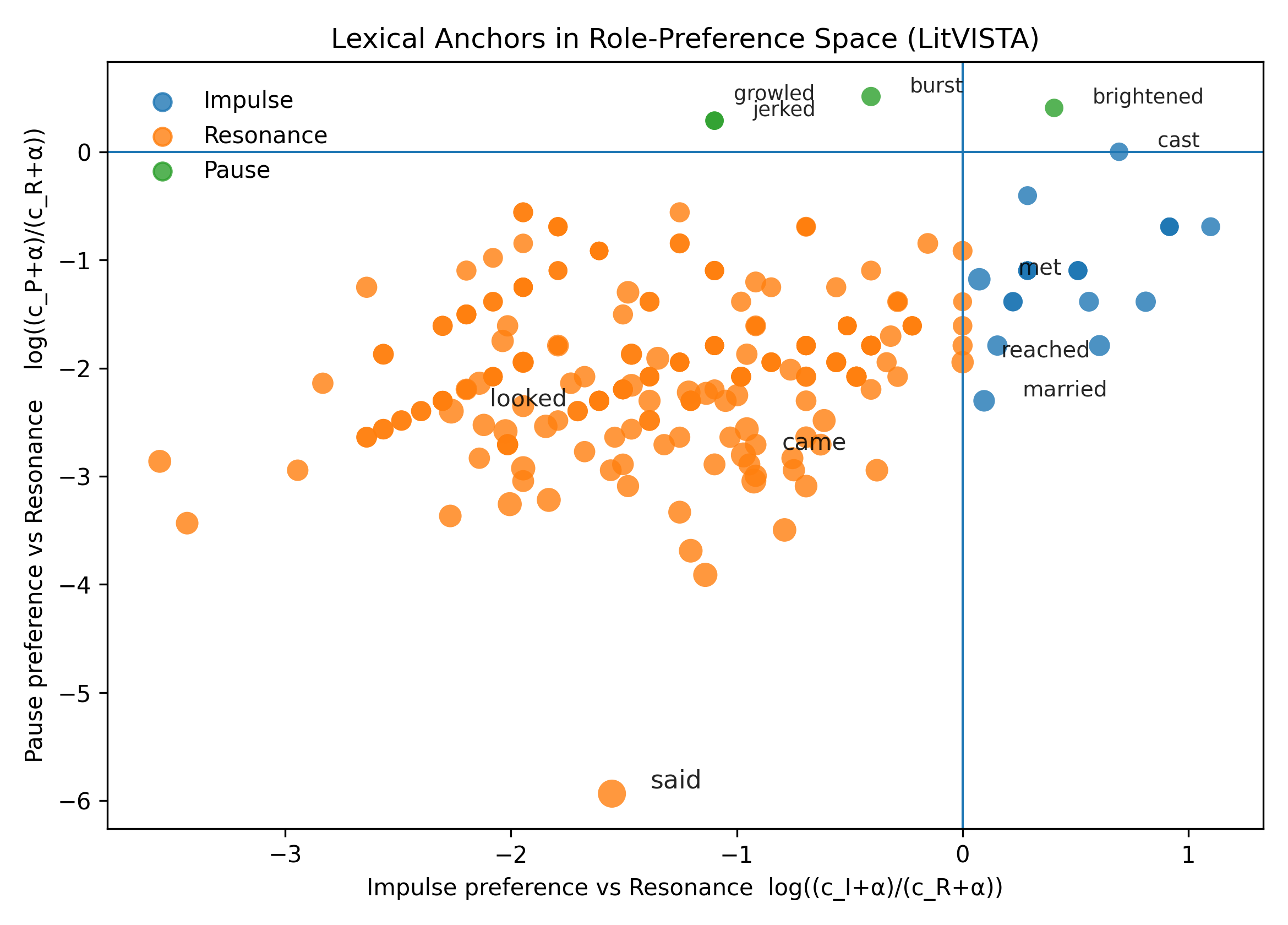}
	\caption{Lexical anchors in role-preference space. The X-axis represents Impulse–Resonance preference, and the Y-axis represents Pause–Resonance preference. Each point corresponds to a lexical item.}
	\label{fig:lexical_roles}
\end{figure}

Figure~\ref{fig:lexical_roles} reveals a structured lexical landscape. Action-oriented verbs such as \textit{cast}, \textit{met}, and \textit{reached} cluster in regions strongly biased toward Impulse, while perception and discourse-related verbs (e.g., \textit{looked}, \textit{said}) occupy Resonance-dominated regions. A smaller set of words aligns with Pause, often corresponding to evaluative or state-descriptive expressions.

Importantly, these clusters emerge without any lexical supervision. The fact that coherent semantic groupings arise purely from narrative role statistics indicates that LitVISTA captures stable associations between lexical items and narrative function. This further supports the claim that the VISTA Space reflects meaningful narrative structure rather than arbitrary annotation artifacts.

\section{Related Work}

Recent work in computational narrative analysis and computational literary studies has shifted from local semantics toward discourse- and structure-level analysis of narrative phenomena, emphasizing plot organization and narrative dynamics in literary texts~\cite{piper-2023-computational}. 
This shift is reinforced by methodological surveys that identify narrative structure as a central object of contemporary computational literary research~\cite{DBLP:journals/it/HatzelSBG23}. 
Related efforts have introduced discourse- and clause-level resources to support large-scale structural analysis of narrative texts~\cite{DBLP:conf/coling/TroianoV24}. 

Event-centric representations remain a common foundation for narrative modeling, with recent work examining how event sequences can be organized into coherent storylines or structured graphs~\cite{DBLP:conf/aaai/VijayaraghavanR23}. 
Other studies investigate narrative consistency by modeling global structural constraints over event sequences rather than isolated relations~\cite{zhu-etal-2023-nlp}. 

In parallel, the rise of frontier large language models has motivated evaluations of narrative understanding on long-form inputs, particularly focusing on long-context and multi-step reasoning~\cite{DBLP:conf/iclr/SpragueYBCD24}. 
Additional work analyzes narrative coherence in generated stories, revealing systematic structural failures despite surface fluency~\cite{zhu-etal-2023-nlp}. 
More recently, evaluations have probed subtext and implicit meaning comprehension in literary narratives~\cite{DBLP:journals/tacl/SubbiahZCM24}. 
At a broader level, structured benchmarks~\cite{DBLP:journals/corr/abs-2503-05244} have also been proposed for evaluating narrative generation and writing quality.

\section{Conclusion}
This paper introduces VISTA Space, a representational framework that unifies human and model perspectives on narrative structure, and LitVISTA, a high-precision benchmark for evaluating narrative orchestration in literary texts. Under oracle evaluation, current frontier language models exhibit persistent difficulty in jointly recovering narrative function and dependency structure, indicating that literary narrative understanding remains fragmented rather than globally integrated. End-to-end analysis further shows that current failures are dominated by upstream anchor identification and localization errors, suggesting that narrative orchestration and event extraction remain only partially coupled in existing systems. We therefore view LitVISTA not as a benchmark for generation itself, but as a diagnostic benchmark for understanding where narrative generation systems still fall short. We hope LitVISTA will support future work on structure-aware narrative analysis, stronger event-grounded modeling, and ultimately more human-like long-form story generation.

\section{Acknowledgements}
This work is supported by the National Natural Science Foundation of China (Grant No. U21B2009).

\clearpage
\section{Limitations}
While LitVISTA serves as a rigorous benchmark for narrative orchestration, we acknowledge several limitations in our current work:

\paragraph{Reliance on Oracle Settings:} Our primary experimental results rely on an oracle setting where candidate event anchors are provided. As discussed in Appendix~\ref{app:end2end}, we found that even frontier LLMs (e.g., GPT-5, Gemini-Pro) currently struggle to perform valid end-to-end narrative reconstruction, primarily due to failures in low-level anchor identification and localization. While this highlights the difficulty of the proposed task, it also limits our current ability to evaluate fully autonomous narrative analysis systems without upstream assistance.

\paragraph{Domain and Language Specificity:} LitVISTA is grounded in the LitBank corpus, which focuses on English literary texts from the public domain. While this choice ensures high-quality, expert-annotated narrative structures and avoids copyright issues, the findings may not fully generalize to other languages, modern internet fiction, or non-literary narrative forms where implicit structural cues might differ.

\paragraph{Annotation Scalability:} To ensure topological consistency and theoretical depth, we employed a resource-intensive expert annotation process with consensus-based adjudication. This high standard for data quality inevitably constrains the scale of our dataset compared to automatically constructed corpora. Consequently, LitVISTA is designed as a high-precision evaluation benchmark rather than a large-scale training corpus.

\paragraph{Subjectivity of Literary Interpretation:} Although we enforce strict axiomatic guidelines (Appendix B) to minimize ambiguity, literary boundaries and structural roles involve inherent interpretative subjectivity. Our "gold standard" represents a coherent, consensus-derived structural reading, but it may not capture every possible valid interpretation of a complex literary passage.

\paragraph{Participant Modeling:} Our current formulation focuses on event-centric structure, modeling narrative orchestration through event anchors and their functional roles. It does not explicitly represent participants or character-level interactions, which are also central to narrative understanding. Extending VISTA to incorporate participant structure remains an important direction for future work.

\section{Ethical Considerations}

\paragraph{Data Source, Licensing, and Privacy: }
The LitVISTA benchmark builds upon the LitBank corpus, a dataset of 100 English-language fiction works sourced from Project Gutenberg. Since these texts belong to the public domain, the dataset contains no personally identifying information (PII) of living individuals. LitBank is licensed under a Creative Commons Attribution 4.0 International License (CC-BY 4.0), and we strictly adhere to these terms in distributing our derived artifacts.

\paragraph{Intended Use: }
Aligning with the scientific intent of Project Gutenberg and LitBank, we release LitVISTA to support research in natural language processing and computational humanities. The benchmark is intended solely for academic research to facilitate the study of narrative dynamics and evaluate the structural capabilities of large language models.

\paragraph{Annotator Compensation and Process: }
The annotation effort involved a small team of contributors; details of annotator background are provided in Section~\ref{anno-background}, while annotation details are described in Section~\ref{anno-detail}. All participants were informed of the research goals and workload in advance, participated on a voluntary basis, and received no monetary compensation.

\paragraph{Use of AI Tools: }
We permitted annotators to use AI tools solely for summarizing broader literary contexts and clarifying plot backgrounds, mitigating the time cost of reading full novels. The core tasks of identifying narrative anchors, assigning topological roles, and resolving dependencies were performed entirely manually by human annotators. No AI-generated labels were used in the construction of the gold standard dataset.

\paragraph{Potential Risks and Subjectivity: }
Literary interpretation involves inherent subjectivity. To mitigate this, we established a multi-phase annotation strategy supported by a Theoretical Codebook (Appendix B) and consensus-based adjudication. While LitVISTA represents a cohesive structural interpretation, users should be aware of the subjective nature characterizing computational literary studies.

\bibliography{custom,anthology-1,anthology-2}
\appendix

\clearpage
\onecolumn

\begin{multicols}{2}
\section{Illustrating Narrative Configuration}
\label{app:narrative_practice}
This appendix provides concrete illustrations of \textit{Narrative Configuration} as defined in Section~\ref{narrative_configuration}. The goal is to clarify how different configurations of the same underlying events give rise to distinct narrative structures through the functional roles of $\mathcal{V}_I$, $\mathcal{V}_R$, and $\mathcal{V}_P$.

Across all examples, the underlying event content remains fixed. What varies is the structural organization imposed by narrative orchestration. These examples demonstrate how narrative meaning emerges from structural configuration rather than from the events themselves.

\subsection{Structural Backbone}
\label{app:skeleton}

At the most basic level, a narrative can be represented as a minimal progression chain composed exclusively of Impulses ($\mathcal{V}_I$). This backbone encodes the irreversible advancement of the narrative state and preserves logical continuity between events.

Consider the following two variants, which share the same set of Impulse events but differ in their ordering:

\begin{description}
	\item[Variation A (Chronological):]
	... Alice \textit{poisons}$_{v_1}$ the coffee ... Bob \textit{drinks}$_{v_2}$ it ... finally ... Bob \textit{is saved}$_{v_3}$ by emergency treatment ...
\end{description}

\begin{description}
	\item[Variation B (Reordered):]
	... Bob \textit{drinks}$_{v_2}$ the coffee ... finally ... Bob \textit{is saved}$_{v_3}$ after a rescue ... the cause is revealed ... Alice had \textit{poisoned}$_{v_1}$ the cup ...
\end{description}

Both variants rely exclusively on $\mathcal{V}_I$ events and therefore encode the same narrative backbone. However, reordering the Impulses alters the distribution of information over narrative time, affecting reader expectation without introducing additional structural operations. This illustrates that even within $\mathcal{V}_I$, narrative effects can arise from configuration rather than content.

\subsection{Lateral Expansion via Resonance}
\label{app:fabric}

While the Impulse chain defines narrative progression, it offers limited expressive capacity. Structural richness emerges when Resonances ($\mathcal{V}_R$) are introduced to laterally expand the narrative state without advancing the progress index.

Using the same Impulse backbone $(\textit{poisons}_{v_1}, \textit{drinks}_{v_2}, \textit{is saved}_{v_3})$, consider the following configuration:

\begin{description}
	\item[Variation C (Resonant Expansion):]
	\uline{Snow \textit{falls}$_{v_R}$ outside while warm jazz \textit{plays}$_{v_R}$.}
	... Bob \textit{drinks}$_{v_2}$ the coffee ... finally ... Bob \textit{is saved}$_{v_3}$ after a rescue ...
\end{description}

Here, the Resonance events attach to the Impulse $\textit{drinks}_{v_2}$, enriching the narrative state without modifying the progression itself. Structurally, $\mathcal{V}_R$ introduces descriptive expansion that shapes reader interpretation while remaining subordinate to the backbone. The resulting narrative effect emerges from the accumulation of contextual information rather than from additional events.

\subsection{Vertical Deepening via Pause}
\label{app:intensity}

Pauses ($\mathcal{V}_P$) operate orthogonally to both progression and expansion. They suspend narrative advancement and concentrate representational density within a single narrative moment.

Consider the following configuration:

\begin{description}
	\item[Variation D (Pause-Induced Density):]
	... Bob \textit{drinks}$_{v_2}$ the coffee,
	\uline{the cup \textit{clatters}$_{v_P}$ to the floor, a high-pitched ring \textit{drowns}$_{v_P}$ out all sound, the ceiling light \textit{stretches}$_{v_P}$ into a star, his heartbeat \textit{slams}$_{v_P}$ to a halt}
	... finally ... Bob \textit{is saved}$_{v_3}$ ...
\end{description}

This sequence of Pause events decomposes a single narrative instant into multiple micro-observations. Rather than advancing the narrative state, these events intensify local representation, producing high expressive density within a fixed temporal window. Structurally, this corresponds to movement along the Z-axis of VISTA Space.

\subsection{Structural Choice and Global Interpretation}
\label{app:intent}

Although Resonances and Pauses are not required to preserve logical continuity, their inclusion determines how the narrative is globally interpreted. Different configurations over the same backbone yield systematically different narrative structures.

The following examples illustrate how discretionary structural choices shape global narrative interpretation:

\begin{description}
	\item[Variation E (Internalization):]
	... Bob \textit{drinks}$_{v_2}$ the coffee ...
	\uline{on the operating table, Bob \textit{recalls}$_{v_P}$ his promise to a dying friend. This memory \textit{ignites}$_{v_R}$ his will to survive}
	... finally ... Bob \textit{is saved}$_{v_3}$ ...
\end{description}

\begin{description}
	\item[Variation F (Externalization):]
	... Bob \textit{drinks}$_{v_2}$ the coffee ...
	\uline{the camera \textit{pans}$_{v_R}$ to a generic logo, then \textit{zooms}$_{v_P}$ in on the brand of the life-support machine}
	... finally ... Bob \textit{is saved}$_{v_3}$ ...
\end{description}

Although both variants preserve the same Impulse structure, their configurations emphasize different narrative dimensions. Variation E concentrates representational mass on internal state transitions, whereas Variation F allocates structural attention to external objects. These differences arise entirely from narrative configuration rather than from changes to event content.

\subsection{Conclusion: Structural Implications for Computation}
\label{app:computational_implications}

These examples demonstrate that narrative meaning is encoded in the structural configuration of events rather than in the events themselves. The Impulse backbone ensures logical progression, while Resonances and Pauses govern expansion and intensification within VISTA Space.

By formalizing these roles and their dependencies, VISTA provides a computationally explicit framework for modeling narrative structure. This framework supports systematic analysis of narrative organization and enables empirical evaluation of whether models construct integrated representations across narrative dimensions.
\end{multicols}

\section{Annotation Guidelines}
\label{app:manual}
We acknowledge the inherent dilemma between minimizing the cognitive load for annotators and maintaining the theoretical depth required for high-complexity tasks. Demanding extensive linguistic expertise is impractical, yet performing topological analysis without theoretical constraints inevitably leads to inconsistency. To resolve this trade-off, we adopted a \textbf{pragmatic tiered strategy}:
\begin{itemize}
	\item The \textbf{Annotator Manual} is designed as the primary, accessible guide for standard workflow, prioritizing intuition over formalism.
	\item The \textbf{Theoretical Codebook} serves as the ultimate axiomatic constitution, intended to be consulted strictly for arbitration during ambiguous or borderline cases.
\end{itemize}

\subsection{VISTA Annotator Manual}
\begin{ManualBox}[VISTA Annotator Manual]
	
	\noindent\textbf{Note to Annotators:} This document outlines the standard operating procedures. For any ambiguity or edge case not covered here, please refer to the Appendix~\ref{app-codebook} for the final axiomatic ruling.
	
	\vspace{0.5em}
	\hrule
	\vspace{1em}
	
	\subsubsection*{1. Task Objective}
	The goal is to reconstruct the linear text into a narrative topology. Annotators must identify \textbf{Narrative Anchors} (verbs) and classify them based on their manipulation of the \textbf{Narrative Progress Index ($\tau$)}.
	
	\subsubsection*{2. Core Classifications}
	Refer to \textbf{Codebook Section 1 \& 2} for formal definitions of $\tau$ and Anchors.
	
	\paragraph{\textcolor{red}{Impulse ($\mathcal{V}_I$)}}
	\begin{itemize}
		\item \textbf{Function:} \textbf{Transition ($\tau \to \tau+1$)}. The story turns the page.
		\item \textbf{The Necessity Test:} Try deleting the verb. If the preceding event cannot logically lead to the subsequent event (creating a causal gap), it is $\mathcal{V}_I$. \textit{(See Codebook Axiom 2.2)}
	\end{itemize}
	
	\paragraph{\textcolor{green}{Resonance ($\mathcal{V}_R$)}}
	\begin{itemize}
		\item \textbf{Function:} \textbf{Micro-shift ($\tau + \epsilon$)}. The story scans the current page.
		\item \textbf{The Texture Test:} If deleting the event removes detail but leaves the logical skeleton intact, it is $\mathcal{V}_R$. \textit{(See Codebook Axiom 3.2)}
	\end{itemize}
	
	\paragraph{\textcolor{blue}{Pause ($\mathcal{V}_P$)}}
	\begin{itemize}
		\item \textbf{Function:} \textbf{Bullet Time ($\tau + 0$)}. The story freezes to gaze deeply.
		\item \textbf{The Density Test:} If a cluster of verbs decomposes a single split-second moment into high-resolution details, it is $\mathcal{V}_P$. \textit{(See Codebook Axiom 4.2)}
	\end{itemize}
	
	\subsubsection*{3. General Principles}
	\begin{itemize}
		\item \textbf{Structure First:} Ignore semantic intensity; focus only on structural function. \textit{(See Codebook Axiom 1.2)}
		\item \textbf{Minimization:} The $\mathcal{V}_I$ chain must be the minimum set required to sustain the plot.
	\end{itemize}
	
	\vspace{1em}
	\hrule
	\vspace{1em}
	
	\subsubsection*{4. Case Study: The Western Duel}
	\begin{quote}
		\textit{Text:} ... The stranger \textbf{draws}$_{[2]}$ his gun. In a flash, he \textbf{pulls}$_{[3]}$ the trigger, the Sheriff \textbf{side-steps}$_{[4]}$, the bullet \textbf{grazes}$_{[5]}$ his hat, the window \textbf{shatters}$_{[6]}$... The Sheriff \textbf{returns}$_{[8]}$ fire...
	\end{quote}
	
	\paragraph{Annotation Workflow Demonstration:}
	
	\textbf{Step 1: Keystone Identification}
	\begin{itemize}
		\item \textbf{draws}$_{[2]}$ and \textbf{returns}$_{[8]}$ are identified as $\mathcal{V}_I$ because they are the minimal nodes required to advance the conflict. \textit{(Refer to Codebook Axiom 6.1)}
	\end{itemize}
	
	\textbf{Step 2: Inertial Filling}
	\begin{itemize}
		\item \textbf{pulls}$_{[3]}$ and \textbf{side-steps}$_{[4]}$ follow the trigger event. By default, they are provisionally marked as $\mathcal{V}_R$ (Accompaniment). \textit{(Refer to Codebook Axiom 6.2)}
	\end{itemize}
	
	\textbf{Step 3: Density Correction}
	\begin{itemize}
		\item \textbf{grazes}$_{[5]}$ and \textbf{shatters}$_{[6]}$ describe micro-physics in a frozen instant.
		\item \textbf{Verdict:} Correct to \textcolor{blue}{$\mathcal{V}_P$}.
		\item \textbf{Reasoning:} These nodes represent a vertical information dive, not a horizontal progression. \textit{(Refer to Codebook Axiom 4.1)}
	\end{itemize}
	
	\subsubsection*{5. Ambiguity Resolution (FAQ)}
	
	\textbf{Q: How to handle psychological actions (thinking, recalling)?}
	\begin{itemize}
		\item \textbf{Verdict:} \textcolor{blue}{$\mathcal{V}_P$ (Pause)}.
		\item \textbf{Reference:} \textbf{Codebook Axiom 4.1}. Internal thoughts are topologically isomorphic to external slow-motion shots; both are vertical dives.
	\end{itemize}
	
	\textbf{Q: How to segment triggers vs. phenomena (e.g., "fired" vs. "sparks")?}
	\begin{itemize}
		\item \textbf{Verdict:} "Fired" is \textcolor{red}{$\mathcal{V}_I$}; "Sparks" is \textcolor{blue}{$\mathcal{V}_P$}.
		\item \textbf{Reference:} \textbf{Codebook Axiom 5.1}. Phenomena are visual residues that must depend on a structural trigger.
	\end{itemize}
	
\end{ManualBox}

\subsection{VISTA Theoretical Codebook}
\label{app-codebook}
This codebook provides the formal foundation of the VISTA annotation scheme, specifying the axiomatic principles that govern role assignment and dependency decisions. It complements the annotator manual by making explicit the theoretical criteria underlying these decisions, particularly in ambiguous or borderline cases.

\begin{ManualBox}[VISTA Theoretical Codebook (Axiomatic System)]	
	\paragraph{1. The Basic Unit Proposition}
	The atom of narrative analysis is the ``Event Operator.''
	\begin{itemize}
		\item \textbf{Axiom 1.1 (Symbolic Proxy):} Verbs are symbolic proxies for underlying semantic units.
		\item \textbf{Axiom 1.2 (The Operator Law):} The value of a verb depends strictly on its \textbf{transformational effect} on the narrative state ($E$), and is orthogonal to its lexical semantic intensity.
	\end{itemize}
	
	\paragraph{2. The Necessity Proposition ($\mathcal{V}_I$)}
	Impulse is the sole logical carrier of narrative progression.
	\begin{itemize}
		\item \textbf{Axiom 2.1 (The Backbone):} $\mathcal{V}_I$ constitutes the irreversible timeline of the story.
		\item \textbf{Axiom 2.2 (Logical Continuity):} Any two adjacent impulses $v_i, v_{i+1}$ must satisfy a direct logical sequence relationship. If $v_i$ is removed, $v_{i+1}$ loses its precondition.
	\end{itemize}
	
	\paragraph{3. The Extension Proposition ($\mathcal{V}_R$)}
	Resonance is the lateral expansion of the narrative dimension.
	\begin{itemize}
		\item \textbf{Axiom 3.1 (Attachment):} $\mathcal{V}_R$ must attach to a backbone node, providing a state description increment ($\delta$).
		\item \textbf{Axiom 3.2 (The Micro-shift):} If $\Delta \text{State} = 0$ (logical index is constant) but physical time flows ($\tau + \epsilon$), the node is $\mathcal{V}_R$.
	\end{itemize}
	
	\paragraph{4. The Depth Proposition ($\mathcal{V}_P$)}
	Pause is the vertical collapse of the narrative dimension.
	\begin{itemize}
		\item \textbf{Axiom 4.1 (Verticality):} $\mathcal{V}_P$ represents a vertical dive into a single moment ($Z$-axis), characterized by high information density and zero narrative velocity ($\tau + 0$).
		\item \textbf{Axiom 4.2 (Super-Resolution):} Any cluster of verbs performing a microscopic decomposition of a single instantaneous frame is defined as $\mathcal{V}_P$.
	\end{itemize}
	
	\paragraph{5. The Structural Proposition}
	\begin{itemize}
		\item \textbf{Axiom 5.1 (Asymmetric Dependency):} All discretionary nodes ($\mathcal{V}_R, \mathcal{V}_P$) must ultimately depend on a structural node ($\mathcal{V}_I$).
		\item \textbf{Axiom 5.1.1 (Ultimate Dependence):} Every discretionary node ultimately resolves to a backbone node ($\mathcal{V}_I$), regardless of local attachment.
		\item \textbf{Axiom 5.1.2 (Operational Serialization):} Intermediate references are operational artifacts, not additional backbone relations.
	\end{itemize}
	
	\paragraph{6. The Operational Proposition}
	Principles for resolving ambiguity during the annotation process.
	\begin{itemize}
		\item \textbf{Axiom 6.1 (Keystone Priority):} The annotation process must prioritize establishing the $\mathcal{V}_I$ chain.
		\item \textbf{Axiom 6.2 (The Relativity Law):} The class of a fuzzy node is determined by its \textbf{axial relationship} relative to the preceding anchor:
		\begin{itemize}
			\item Progression $\to \mathcal{V}_I$
			\item Accompaniment $\to \mathcal{V}_R$
			\item Deepening $\to \mathcal{V}_P$
		\end{itemize}
	\end{itemize}
	
\end{ManualBox}

\section{Concrete Annotated Example}
\label{app:example}

\paragraph{Visual Representation Note:}
In the actual VISTA dataset, topological labels are encoded using inline HTML-style tags (e.g., \texttt{<span style=``color:red''>verb</span>}). This encoding scheme is a deliberate design choice, calculated to leverage the inherent proficiency of modern Large Language Models (LLMs) in handling structured formatting constraints (e.g., HTML/XML schemas), thereby enhancing topological consistency during generation. 

For the sake of readability in this document, we have rendered these raw tags directly as colored text. The color coding and notation scheme are defined as follows:
\begin{itemize}
	\item \textbf{\textcolor{vRed}{Red}}: Impulse ($\mathcal{V}_I$), denoting narrative progression.
	\item \textbf{\textcolor{vGreen}{Green}}: Resonance ($\mathcal{V}_R$), denoting descriptive expansion.
	\item \textbf{\textcolor{vBlue}{Blue}}: Pause ($\mathcal{V}_P$), denoting vertical deepening.
	\item \textbf{Indices (@n / \#n)}: Indicate the topological dependency between a governing node (@) and its dependent (\#).
\end{itemize}

Below is an excerpt of an annotation sample from \textit{Alice's Adventures in Wonderland}.
\subsubsection*{Input (Raw Text)}
\begin{DataBox}
[Excerpt from \textit{Alice's Adventures in Wonderland}]
\textbf{Chapter I. Down the Rabbit-Hole}

Alice was beginning to get very tired of sitting by her sister on the bank, and of having nothing to do: once or twice she had peeped into the book her sister was reading, but it had no pictures or conversations in it, ``and what is the use of a book,'' thought Alice ``without pictures or conversations?''

So she was considering in her own mind (as well as she could, for the hot day made her feel very sleepy and stupid), whether the pleasure of making a daisy-chain would be worth the trouble of getting up and picking the daisies, when suddenly a White Rabbit with pink eyes ran close by her.

There was nothing so \textit{very} remarkable in that; nor did Alice think it so \textit{very} much out of the way to hear the Rabbit say to itself, ``Oh dear! Oh dear! I shall be late!'' (when she thought it over afterwards, it occurred to her that she ought to have wondered at this, but at the time it all seemed quite natural); but when the Rabbit actually \textsc{Took a Watch Out of Its Waistcoat-Pocket}, and looked at it, and then hurried on, Alice started to her feet, for it flashed across her mind that she had never before seen a rabbit with either a waistcoat-pocket, or a watch to take out of it, and burning with curiosity, she ran across the field after it, and fortunately was just in time to see it pop down a large rabbit-hole under the hedge.

In another moment down went Alice after it, never once considering how in the world she was to get out again. The rabbit-hole went straight on like a tunnel for some way, and then dipped suddenly down, so suddenly that Alice had not a moment to think about stopping herself before she found herself falling down a very deep well.

Either the well was very deep, or she fell very slowly, for she had plenty of time as she went down to look about her and to wonder what was going to happen next. First, she tried to look down and make out what she was coming to, but it was too dark to see anything; then she looked at the sides of the well, and noticed that they were filled with cupboards and book-shelves; here and there she saw maps and pictures hung upon pegs. She took down a jar from one of the shelves as she passed; it was labelled `\textsc{Orange Marmalade}', but to her great disappointment it was empty: she did not like to drop the jar for fear of killing somebody, so managed to put it into one of the cupboards as she fell past it.

``Well!'' thought Alice to herself, ``after such a fall as this, I shall think nothing of tumbling down stairs! How brave they'll all think me at home! Why, I wouldn't say anything about it, even if I fell off the top of the house!'' (Which was very likely true.)

\end{DataBox}

\subsubsection*{Output (Topological Annotation)}
\begin{DataBox}
Down the Rabbit-Hole Alice was beginning to get very \imp{tired@1} of sitting by her sister on the bank, and of having nothing to do: once or twice she had \res{peeped\#1} into the book her sister was \res{reading}, but it had no pictures or conversations in it, ``and what is the use of a book,'' \pau{thought} Alice ``without pictures or conversations?''

So she was \pau{considering} in her own mind (as well as she could, for the hot day made her feel very sleepy and stupid), whether the pleasure of making a daisy-chain would be worth the trouble of getting up and picking the daisies, when suddenly a White Rabbit with pink eyes \imp{ran} close by her.

There was nothing so \textsc{very} remarkable in that; nor did Alice think it so \textsc{very} much out of the way to \imp{hear@2} the Rabbit \imp{say} to itself, ``Oh dear! Oh dear! I shall be late!'' (when she \res{thought\#2} it over afterwards, it \res{occurred} to her that she ought to have wondered at this, but at the time it all seemed quite natural); but when the Rabbit actually \res{TOOK} A WATCH OUT OF ITS WAISTCOAT-POCKET, and \res{looked} at it, and then \res{hurried} on, Alice \imp{started} to her feet, for it \imp{flashed} across her mind that she had never before seen a rabbit with either a waistcoat-pocket, or a watch to take out of it, and burning with curiosity, she \imp{ran} across the field after it, and fortunately was just in time to \imp{see} it \imp{pop} down a large rabbit-hole under the hedge.

In another moment down \imp{went} Alice after it, never once considering how in the world she was to get out again. The rabbit-hole went straight on like a tunnel for some way, and then dipped suddenly down, so suddenly that Alice had not a moment to think about stopping herself before she found herself \imp{falling} down a very deep well.

Either the well was very deep, or she fell very slowly, for she had plenty of time as she \imp{went} down to \imp{look} about her and to \imp{wonder} what was going to happen next. First, she \imp{tried} to look down and make out what she was coming to, but it was too dark to see anything; then she \imp{looked} at the sides of the well, and \imp{noticed@3} that they were filled with cupboards and book-shelves; here and there she saw maps and pictures hung upon pegs.

She \res{took\#3} down a jar from one of the shelves as she \res{passed}; it was labelled ``\textsc{Orange Marmalade}'', but to her great \res{disappointment} it was empty: she did not like to drop the jar for fear of killing somebody, so managed to \res{put} it into one of the cupboards as she \res{fell} past it.

``Well!'' \imp{thought} Alice to herself, ``after such a \imp{fall} as this, I shall think nothing of tumbling down stairs! How brave they'll all think me at home! Why, I wouldn't say anything about it, even if I fell off the top of the house!'' (Which was very likely true.)

\end{DataBox}

\section{Details of Experimental Settings}
\label{app:exp_setup}
This section summarizes the experimental settings and prompt designs. We set the temperature to \texttt{0.0} whenever applicable; otherwise, default settings are used.

We use two prompt variants. The Oracle Evaluation Prompt (Appendix~\ref{app:oracle_prompt}) takes the raw text together with event anchors and their character offsets, whereas the End-to-End Evaluation Prompt (Appendix~\ref{app:e2e_prompt}) takes only the raw text and therefore specifies anchor definitions more explicitly. Both prompts include one fully annotated example.

%

\subsection{Oracle Evaluation Prompt}
\label{app:oracle_prompt}
\begin{DataBox}
	\textbf{Prompt: Narrative Topology Classification (Pre-identified Anchors)}
	
	\vspace{0.5em}
	\noindent\textbf{System Instruction:} 
	You are an expert Narrative Analyst. You are tasked with analyzing a text to construct a structured dependency graph.
	
	\vspace{0.5em}
	\noindent\textbf{CRITICAL CHANGE:} You do NOT need to extract words from scratch. You will be provided with the \textbf{Input Text} and a list of \textbf{Pre-identified Anchors} (comprising ID, Offsets, and Word).
	
	Your task is to assign the correct \textbf{Category} and \textbf{Head} (dependency) for each provided Anchor, strictly following the framework definitions below.
	
	\vspace{0.5em}
	\noindent\textbf{I. Foundational Definitions}
	
	\textbf{The Narrative Anchor ($v$)}
	An Anchor is a symbolic proxy for a semantic event or state change.
	\begin{itemize}
		\item \textbf{Context:} You are provided with these Anchors. They include Finite Verbs (e.g., ``draws'') and Event Nominals (e.g., ``departure'').
		\item \textbf{Your Job:} Do not add or remove anchors. Analyze only the ones provided in the list.
	\end{itemize}
	
	\textbf{The Narrative Progress Index ($\tau$)}
	Narrative time is NOT chronological time. We track the Narrative Progress Index ($\tau$), which represents the logical stage of the plot.
	\begin{itemize}
		\item \textbf{Rule:} $\tau$ only increments when the narrative state \textit{must} change to enable the next event.
		\item \textbf{Constraint:} Mere descriptions or internal thoughts do not advance $\tau$; they expand the current stage.
	\end{itemize}
	
	\vspace{0.5em}
	\noindent\textbf{II. Task Definitions: The Topological Roles}
	
	For every provided \textbf{Anchor}, you must classify its operation on the Index ($\tau$) using the following three roles:
	
	\textbf{Role A: IMPULSE (The Plot Driver)}
	\begin{itemize}
		\item \textbf{Operation:} $\tau + 1$ (Advances the Index).
		\item \textbf{Definition:} These are the backbone events. They irreversibly change the state of the story.
		\item \textbf{The Necessity Test:} If you delete this anchor, does the logical chain break? If the next event loses its cause/precondition, this is an Impulse.
	\end{itemize}
	
	\textbf{Role B: RESONANCE (The Lateral Expansion)}
	\begin{itemize}
		\item \textbf{Operation:} $\tau$ (Same Index, Lateral shift).
		\item \textbf{Definition:} These events happen alongside the Impulse to provide atmosphere, manner, or context.
		\item \textbf{The Texture Test:} If you delete this anchor, is the plot skeleton preserved, losing only descriptive detail? If yes, it is a Resonance.
	\end{itemize}
	
	\textbf{Role C: PAUSE (The Vertical Intensity)}
	\begin{itemize}
		\item \textbf{Operation:} $\tau$ (Index Freeze).
		\item \textbf{Definition:} The narrative flow halts to load ``Information Density'' into a single moment.
		\item \textbf{The Density Test:} Does this anchor represent a split-second micro-action (physics) or a dive into internal psychology (thoughts)? If it dives ``inward'' instead of moving ``forward,'' it is a Pause.
	\end{itemize}
	
	\vspace{0.5em}
	\noindent\textbf{III. Dependency Logic (Determining the ``Head'')}
	
	\begin{itemize}
		\item \textbf{If Impulse:} Points to the \textbf{previous Impulse} ID (or -1 if it is the first/root).
		\item \textbf{If Resonance/Pause:} Points to the ID of the \textbf{Impulse} that governs the current state (the Impulse being modified or described).
	\end{itemize}
	
	\vspace{0.5em}
	\noindent\textbf{IV. Output Formatting Strategy}
	
	You must output a structured list (simulated table).
	\begin{itemize}
		\item \textbf{Format:} Tab-separated or fixed-width text.
		\item \textbf{Constraint:} The \texttt{ID}, \texttt{Offsets}, and \texttt{Word} columns must MATCH the Input Anchors exactly.
	\end{itemize}
	
	\textbf{Columns Definition:}
	\begin{enumerate}
		\item \textbf{ID}: The unique integer provided in the input.
		\item \textbf{Category}: Your classification (Impulse, Resonance, or Pause).
		\item \textbf{Offsets}: The offsets provided in the input (e.g., \texttt{331,334}).
		\item \textbf{Word}: The word provided in the input.
		\item \textbf{Head}: The ID of the parent node (calculated by you).
	\end{enumerate}
	
	\textbf{Output Template:}
	\begin{verbatim}
		ID   Category     Offsets     Word     Head
		0    Resonance    331,334     had      1
		1    Impulse      796,803     imputes  -1
	\end{verbatim}
	
	\vspace{0.5em}
	\noindent\textbf{V. One-Shot Demonstration}
	
	\textbf{Input Text:} ``CHAPTER I. Down the Rabbit-Hole Alice was beginning to get very tired of sitting by her sister on the bank , and of having nothing to do : once or twice she had peeped into the book her sister was reading , but it had no pictures or conversations in it , ` and what is the use of a book , ' thought Alice ` without pictures or conversations ? '
	So she was considering in her own mind ( as well as she could , for the hot day made her feel very sleepy and stupid ) , whether the pleasure of making a daisy-chain would be worth the trouble of getting up and picking the daisies , when suddenly a White Rabbit with pink eyes ran close by her .
	There was nothing so VERY remarkable in that ; nor did Alice think it so VERY much out of the way to hear the Rabbit say to itself , ` Oh dear !
	Oh dear !
	I shall be late ! '
	( when she thought it over afterwards , it occurred to her that she ought to have wondered at this , but at the time it all seemed quite natural ) ; but when the Rabbit actually TOOK A WATCH OUT OF ITS WAISTCOAT-POCKET , and looked at it , and then hurried on , Alice started to her feet , for it flashed across her mind that she had never before seen a rabbit with either a waistcoat-pocket , or a watch to take out of it , and burning with curiosity , she ran across the field after it , and fortunately was just in time to see it pop down a large rabbit-hole under the hedge .''
	
	\textbf{Input Anchors:}
	\begin{verbatim}
		0    64,69       tired
		1    161,167     peeped
		2    197,204     reading
		3    291,298     thought
		4    356,367     considering
		5    622,625     ran
		6    742,746     hear
		7    758,761     say
		8    827,834     thought
		9    859,867     occurred
		10   994,998     TOOK
		11   1041,1047   looked
		12   1065,1072   hurried
		13   1084,1091   started
		14   1113,1120   flashed
		15   1274,1277   ran
		16   1342,1345   see
		17   1349,1352   pop
	\end{verbatim}
	
	\textbf{Reasoning:}
	\begin{enumerate}
		\item tired (ID 0): State change (becoming tired). Advances narrative state. $\rightarrow$ Impulse. Head: -1.
		\item peeped (ID 1): Minor action occurring alongside the main state. Does not advance plot stage. $\rightarrow$ Resonance. Head: 0.
		\item reading (ID 2): Contextual activity of the sister. Expands the scene. $\rightarrow$ Resonance. Head: 1.
		\item thought (ID 3): Internal mental process. Freezes time to load information. $\rightarrow$ Pause. Head: 0.
	\end{enumerate}
	
	\textbf{Output:}
	\begin{verbatim}
		0    Impulse     64,69       tired       -1
		1    Resonance   161,167     peeped      0
		2    Resonance   197,204     reading     1
		3    Pause       291,298     thought     0
		4    Pause       356,367     considering 3
		5    Impulse     622,625     ran         0
		6    Impulse     742,746     hear        5
		7    Impulse     758,761     say         6
		8    Resonance   827,834     thought     6
		9    Resonance   859,867     occurred    8
		10   Resonance   994,998     TOOK        9
		11   Resonance   1041,1047   looked      10
		12   Resonance   1065,1072   hurried     11
		13   Impulse     1084,1091   started     7
		14   Impulse     1113,1120   flashed     13
		15   Impulse     1274,1277   ran         14
		16   Impulse     1342,1345   see         15
		17   Impulse     1349,1352   pop         16
		Any other text is prohibited from being output.
	\end{verbatim}
	
	\vspace{0.5em}
	\noindent\textbf{VI. Task}
	
	\textbf{Input Text:} [INSERT TEXT HERE]
	
	\textbf{Input Anchors:} [INSERT ANCHOR LIST HERE (Format: ID  Offsets  Word)]
\end{DataBox}

\subsection{End-to-End Evaluation Prompt} 
\label{app:e2e_prompt}
\begin{DataBox}
	\textbf{System Instruction:} 
	You are an expert Narrative Analyst. You are tasked with deconstructing a text into a structured dependency graph. To do this, you must first understand the fundamental definitions of the framework provided below. Do not rely on outside knowledge; strictly follow these definitions.
	
	\vspace{0.5em}
	\noindent\textbf{I. Foundational Definitions}
	
	\textbf{The Narrative Anchor ($v$)}
	Before analyzing structure, you must identify the atomic units of the narrative, called Anchors.
	\begin{itemize}
		\item \textbf{Definition:} An Anchor is a symbolic proxy for a semantic event or state change.
		\item \textbf{Scope:} This includes \textbf{Finite Verbs} (e.g., ``draws'', ``ran'') AND \textbf{Event Nominals} (nouns that imply an event structure, e.g., ``departure'', ``marriage'', ``thought'').
		\item \textbf{Exclusion:} Do NOT tag auxiliary verbs (is, was, had) or functional connectors unless they are the sole carrier of meaning.
	\end{itemize}
	
	\textbf{The Narrative Progress Index ($\tau$)}
	Narrative time is NOT chronological time. We track the Narrative Progress Index ($\tau$), which represents the logical stage of the plot.
	\begin{itemize}
		\item \textbf{Rule:} $\tau$ only increments when the narrative state \textit{must} change to enable the next event.
		\item \textbf{Constraint:} Mere descriptions or internal thoughts do not advance $\tau$; they expand the current stage.
	\end{itemize}
	
	\vspace{0.5em}
	\noindent\textbf{II. Task Definitions: The Topological Roles}
	
	For every identified \textbf{Anchor}, you must classify its operation on the Index ($\tau$) using the following three roles:
	
	\textbf{Role A: IMPULSE (The Plot Driver)}
	\begin{itemize}
		\item \textbf{Operation:} $\tau + 1$ (Advances the Index).
		\item \textbf{Definition:} These are the backbone events. They irreversibly change the state of the story.
		\item \textbf{The Necessity Test:} If you delete this anchor, does the logical chain break? If the next event loses its cause/precondition, this is an Impulse.
	\end{itemize}
	
	\textbf{Role B: RESONANCE (The Lateral Expansion)}
	\begin{itemize}
		\item \textbf{Operation:} $\tau$ (Same Index, Lateral shift).
		\item \textbf{Definition:} These events happen alongside the Impulse to provide atmosphere, manner, or context.
		\item \textbf{The Texture Test:} If you delete this anchor, is the plot skeleton preserved, losing only descriptive detail? If yes, it is a Resonance.
	\end{itemize}
	
	\textbf{Role C: PAUSE (The Vertical Intensity)}
	\begin{itemize}
		\item \textbf{Operation:} $\tau$ (Index Freeze).
		\item \textbf{Definition:} The narrative flow halts to load ``Information Density'' into a single moment.
		\item \textbf{The Density Test:} Does this anchor represent a split-second micro-action (physics) or a dive into internal psychology (thoughts)? If it dives ``inward'' instead of moving ``forward,'' it is a Pause.
	\end{itemize}
	
	\vspace{0.5em}
	\noindent\textbf{III. Output Formatting Strategy}
	
	You must output the analysis as a structured list (simulated table) containing the following columns. Do NOT use HTML tags.
	
	\textbf{Columns Definition:}
	\begin{enumerate}
		\item \textbf{ID}: A unique sequential integer (0, 1, 2...) for each Anchor found.
		\item \textbf{Category}: The Role (Impulse, Resonance, or Pause).
		\item \textbf{Offsets}: The start and end character position of the word in the input text (e.g., \texttt{331,334}). \textit{Note: Estimate the offsets as accurately as possible based on the provided text.}
		\item \textbf{Word}: The exact text of the Anchor.
		\item \textbf{Head}: The ID of the parent node.
		\begin{itemize}
			\item \textbf{If Impulse}: Points to the \textit{previous} Impulse ID (or -1 if it is the first/root).
			\item \textbf{If Resonance/Pause}: Points to the ID of the \textbf{Impulse} that governs the current state (the Impulse being modified).
		\end{itemize}
	\end{enumerate}
	
	\textbf{Output Template:}
	\begin{verbatim}
		ID   Category     Offsets     Word     Head
		0    Resonance    331,334     had      1
		1    Impulse      796,803     imputes  -1
	\end{verbatim}
	
	\vspace{0.5em}
	\noindent\textbf{IV. One-Shot Demonstration}
	
	\textbf{Input Text:} ``CHAPTER I. Down the Rabbit-Hole Alice was beginning to get very tired of sitting by her sister on the bank , and of having nothing to do : once or twice she had peeped into the book her sister was reading , but it had no pictures or conversations in it , ` and what is the use of a book , ' thought Alice ` without pictures or conversations ? '
	So she was considering in her own mind ( as well as she could , for the hot day made her feel very sleepy and stupid ) , whether the pleasure of making a daisy-chain would be worth the trouble of getting up and picking the daisies , when suddenly a White Rabbit with pink eyes ran close by her .
	There was nothing so VERY remarkable in that ; nor did Alice think it so VERY much out of the way to hear the Rabbit say to itself , ` Oh dear !
	Oh dear !
	I shall be late ! '
	( when she thought it over afterwards , it occurred to her that she ought to have wondered at this , but at the time it all seemed quite natural ) ; but when the Rabbit actually TOOK A WATCH OUT OF ITS WAISTCOAT-POCKET , and looked at it , and then hurried on , Alice started to her feet , for it flashed across her mind that she had never before seen a rabbit with either a waistcoat-pocket , or a watch to take out of it , and burning with curiosity , she ran across the field after it , and fortunately was just in time to see it pop down a large rabbit-hole under the hedge .''
	
	\textbf{Reasoning:}
	\begin{enumerate}
		\item \textbf{tired} (ID 0): State change (becoming tired). Advances narrative state. $\rightarrow$ \textbf{Impulse}. Head: -1.
		\item \textbf{peeped} (ID 1): Minor action occurring alongside the main state. Does not advance plot stage. $\rightarrow$ \textbf{Resonance}. Head: 0.
		\item \textbf{reading} (ID 2): Contextual activity of the sister. Expands the scene. $\rightarrow$ \textbf{Resonance}. Head: 1.
		\item \textbf{thought} (ID 3): Internal mental process. Freezes time to load information. $\rightarrow$ \textbf{Pause}. Head: 0.
	\end{enumerate}
	
	\textbf{Output:}
	\begin{verbatim}
		0    Impulse     64,69       tired       -1
		1    Resonance   161,167     peeped      0
		2    Resonance   197,204     reading     1
		3    Pause       291,298     thought     0
		4    Pause       356,367     considering 3
		5    Impulse     622,625     ran         0
		6    Impulse     742,746     hear        5
		7    Impulse     758,761     say         6
		8    Resonance   827,834     thought     6
		9    Resonance   859,867     occurred    8
		10   Resonance   994,998     TOOK        9
		11   Resonance   1041,1047   looked      10
		12   Resonance   1065,1072   hurried     11
		13   Impulse     1084,1091   started     7
		14   Impulse     1113,1120   flashed     13
		15   Impulse     1274,1277   ran         14
		16   Impulse     1342,1345   see         15
		17   Impulse     1349,1352   pop         16
	\end{verbatim}
	
	\vspace{0.5em}
	\noindent\textbf{V. Task}
	
	Analyze the following text strictly following the Definitions, Logical Tests, and Output Format above.
	
	\textbf{Input Text:} [INSERT TEXT HERE]
\end{DataBox}

\begin{multicols}{2}
\section{End-to-End Analysis}
\label{app:end2end}
This appendix presents an end-to-end analysis to complement the oracle event-level experiments reported in the main paper. The goal of this analysis is to examine whether current large language models can perform narrative orchestration when provided only with raw text and a fully specified prompt, without access to gold event anchors.

We first summarize the overall findings and failure modes observed in the end-to-end setting (Section~\ref{app:end2end_results}). We then present representative model outputs alongside the corresponding ground-truth annotations to illustrate the observed errors in detail (Section~\ref{app:end2end_examples}).

\subsection{End-to-End Results and Analysis}
\label{app:end2end_results}

We evaluate a representative set of frontier models in an end-to-end setting, including DeepSeek-v3.2, Gemini-3-Pro-Preview-Thinking, GPT-5, GPT-5-Thinking, Grok-4.1-Thinking, and Qwen3-235B-A22. In this setting, models are provided only with the raw narrative text and a fully specified prompt that defines narrative anchors, their functional roles, and the dependency structure, along with a concrete illustrative example.

Across all tested models, performance in the end-to-end setting is uniformly zero. Specifically, none of the models are able to produce a valid reconstruction of the LitVISTA graph that satisfies the evaluation criteria.

To diagnose the source of this failure, we analyze the raw model outputs in detail. Representative predictions are shown in Section~\ref{app:end2end_examples} alongside the corresponding ground-truth annotations. Two systematic failure modes consistently emerge:

\begin{itemize}
	\item \textbf{Incomplete anchor identification}: Given a narrative with around one hundred events, a substantial fraction of anchors are consistently omitted. Models fail to exhaustively identify all event anchors in the text. For example, in the case of DeepSeek-v3.2, numerous event anchors like "CONTAINING" and "BIRTH" appear, but several key events like "lived" and "proceed" are omitted.
	\item \textbf{Misalignment of spans}: Even when an anchor is identified, models often mis-specify its exact token span or positional offset, leading to misaligned or invalid anchors. For instance, GPT-5 outputs anchors such as "CONDESCENDED" but misaligns spans (e.g., "2500,2511") that don't correspond to the actual ground-truth position.
\end{itemize}

These errors are characteristic of probabilistic, generative models. Exhaustive anchor extraction and precise span localization require strict coverage guarantees and exact alignment with the source text, properties that current autoregressive generation paradigms do not reliably provide. Because anchor identification and localization constitute the first step in the narrative reconstruction pipeline, errors at this stage prevent subsequent role assignment and dependency resolution from being meaningfully evaluated, resulting in zero scores under the end-to-end setting.

Taken together, these results indicate that the observed end-to-end failure reflects limitations in upstream anchor identification and localization rather than deficiencies in model capacity or dataset quality. As demonstrated in the main paper under the oracle setting, multiple models achieve strong performance when gold event anchors are provided. For example, Claude-sonnet-4-thinking attains a balanced Anchor F1 of 0.4914 and a Dependency F1 of 0.5624, while GPT-5.1-thinking reaches a Dependency Parsing F1 as high as 0.8135. These findings confirm that the downstream narrative orchestration task itself is well within the representational capacity of current models.

\subsection{Representative Model Outputs}
\label{app:end2end_examples}

To qualitatively illustrate the failure modes discussed above, we present representative end-to-end predictions produced by different models on the same narrative input. The example is drawn from a single chapter of \textit{The History of Tom Jones, a Foundling}, for which the LitVISTA annotation contains exactly fourteen event anchors.

For each model, we report its predicted anchors together with assigned roles, token offsets, and dependency heads. While the gold annotation consists of a compact and well-defined set of anchors, model predictions typically contain substantially more entries, along with omissions, misaligned spans, and structural inconsistencies. Ellipses indicate omitted portions of the prediction.
\end{multicols}

\paragraph{Ground Truth (LitVISTA Annotation).}
The gold annotation contains exactly fourteen event anchors. All anchors are shown in full below.

\begin{DataBox}
	\begin{tabular}{ccccc}
		ID & Category & Offsets & Word & Head \\
		0  & Impulse   & 2500,2511  & condescended  & -1  \\
		1  & Resonance & 2650,2655  & prefix        & 0   \\
		2  & Resonance & 2750,2753  & give          & 0   \\
		3  & Resonance & 2900,2903  & made          & 0   \\
		4  & Pause     & 3200,3203  & fear          & 0   \\
		5  & Resonance & 3600,3608  & represent     & 0   \\
		6  & Resonance & 3720,3723  & hash          & 0   \\
		7  & Pause     & 3950,3954  & doubt         & 0   \\
		8  & Resonance & 4250,4255  & detain        & 0   \\
		9  & Impulse   & 4330,4336  & proceed       & 0   \\
		10 & Impulse   & 4700,4704  & lived         & 9   \\
		11 & Resonance & 4780,4785  & called        & 10  \\
		12 & Resonance & 4920,4928  & contended     & 10  \\
		13 & Resonance & 5070,5077  & bestowed      & 10  \\
		14 & Resonance & 5420,5426  & decreed       & 10  \\
	\end{tabular}
\end{DataBox}

\paragraph{GPT-5.}
GPT-5 generates fewer anchors than DeepSeek-v3.2, but still exceeds the gold count and fails to recover the complete gold structure.

\begin{DataBox}
\begin{tabular}{ccccc}
	ID & Category & Offsets & Word & Head \\
	0  & Impulse   & 1150,1162  & condescended  & -1  \\
	1  & Resonance & 1177,1183  & prefix        & 0   \\
	2  & Resonance & 1217,1220  & give          & 0   \\
	3  & Resonance & 1310,1314  & made          & 0   \\
	4  & Impulse   & 2045,2052  & proceed       & 0   \\
	5  & Impulse   & 2210,2215  & lived         & 4   \\
	6  & Resonance & 2635,2642  & married       & 5   \\
	7  & Impulse   & 2700,2704  & died          & 5   \\
	8  & Impulse   & 2748,2755  & burying       & 7   \\
	9  & Pause     & 2860,2863  & said          & 8   \\
	10 & Pause     & 2888,2897  & considered    & 8   \\
	11 & Impulse   & 2938,2943  & lived         & 8   \\
	12 & Resonance & 3245,3251  & intend        & 11  \\
	13 & Resonance & 3380,3383  & told          & 11  \\
	14 & Resonance & 3412,3421  & inherited     & 11  \\
	15 & Resonance & 3610,3615  & built         & 11  \\
\end{tabular}
\end{DataBox}

\paragraph{GPT-5-Thinking.}
GPT-5-Thinking generates a sequence of event anchors, though it still produces errors in coverage, span localization, and anchor alignment. Below, we show the full output for the first 17 predicted anchors, followed by the last two anchors.

\begin{DataBox}
	\begin{tabular}{ccccc}
		ID & Category & Offsets & Word & Head \\
		0  & Impulse   & 131,144  & ought       & -1  \\
		1  & Resonance & 157,162  & consider    & 0   \\
		2  & Resonance & 253,263  & provides    & 0   \\
		3  & Resonance & 331,345  & is          & 0   \\
		4  & Impulse   & 347,353  & may         & 0   \\
		5  & Resonance & 490,502  & known       & 4   \\
		6  & Resonance & 550,566  & have        & 0   \\
		7  & Impulse   & 573,586  & ensure      & 4   \\
		\multicolumn{5}{c}{\textbf{$\cdots$}} \\
		18 & Impulse   & 1240,1246  & fear       & 16  \\
		19 & Resonance & 1315,1326  & very       & 17  \\
	\end{tabular}
\end{DataBox}

\paragraph{DeepSeek-v3.2.}
DeepSeek-v3.2 produces a long sequence of predicted anchors that substantially exceeds the fourteen gold events. Below we show the beginning of the prediction in full, followed by selected later segments.

\begin{DataBox}
	\begin{tabular}{ccccc}
		ID & Category & Offsets & Word & Head \\
		0  & Impulse   & 7,15     & CONTAINING   & -1  \\
		1  & Resonance & 27,34    & BIRTH        & 0   \\
		2  & Resonance & 81,90    & ACQUAINT     & 0   \\
		3  & Impulse   & 212,224  & consider     & 0   \\
		4  & Resonance & 282,287  & gives        & 3   \\
		5  & Resonance & 373,379  & keeps        & 3   \\
		6  & Resonance & 455,464  & provides     & 5   \\
		7  & Resonance & 502,511  & disagreeable  & 6   \\
		\multicolumn{5}{c}{\textbf{$\cdots$}} \\
		122 & Impulse  & 9737,9742 & plead       & 118 \\
		123 & Resonance & 9983,9995 & inherited  & 122 \\
	\end{tabular}
\end{DataBox}

\paragraph{Gemini-3-Pro-Preview-Thinking.}
Gemini-3-Pro-Preview-Thinking produces the sparsest output among the models shown, yet still fails to recover all fourteen gold anchors.

\begin{DataBox}
	\begin{tabular}{ccccc}
		ID & Category & Offsets & Word & Head \\
		0  & Impulse   & 183,191  & consider    & -1  \\
		1  & Resonance & 227,232  & gives       & 0   \\
		2  & Resonance & 283,288  & keeps       & 0   \\
		3  & Impulse   & 612,619  & happens     & 0   \\
		4  & Resonance & 678,684  & insist      & 3   \\
		5  & Impulse   & 868,875  & prevent     & 3   \\
		6  & Impulse   & 1317,1329 & condescended & 5   \\
		7  & Impulse   & 1392,1398 & prefix      & 6   \\
		\multicolumn{5}{c}{\textbf{$\cdots$}} \\
		28 & Impulse   & 6692,6696 & told        & 27  \\
		29 & Impulse   & 6813,6822 & concluded   & 28  \\
	\end{tabular}
\end{DataBox}

\end{document}